\documentclass[twoside]{article}

\usepackage{aistats2020}

\usepackage[hidelinks]{hyperref}
\usepackage{graphicx}
\usepackage{url}
\usepackage{xcolor}
\usepackage{amsmath}
\usepackage{amsfonts}
\usepackage{amssymb}
\usepackage{subfig}
\usepackage{array}
\usepackage{bm}
\usepackage{booktabs}

\newcommand{\mbf}[1]{\mathbf{#1}}
\newcommand{\mcl}[1]{\mathcal{#1}}
\newcommand{\mbb}[1]{\mathbb{#1}}
\newcommand{\argmax}{\text{arg}\!\max}

\newtheorem{theorem}{Theorem}

\begin{document}

\twocolumn[

\aistatstitle{Rectified Max-Value Entropy Search for Bayesian Optimization}

\aistatsauthor{Quoc Phong Nguyen\textsuperscript{1}, Bryan Kian Hsiang Low\textsuperscript{1}, Patrick Jaillet\textsuperscript{2} }

\aistatsaddress{ 
\textsuperscript{1}Dept. of Computer Science,
National University of Singapore,
Repulic of Singapore\\
\texttt{\{qphong, lowkh\}@comp.nus.edu.sg}\\
 \textsuperscript{2}Dept. of Electrical Engineering and Computer Science, MIT, USA\\
\texttt{jaillet@mit.edu}}
]

\begin{abstract}
Although the existing \emph{max-value entropy search} (MES) is based on the widely celebrated notion of mutual information, its empirical performance can suffer due to two misconceptions whose implications on the exploration-exploitation trade-off are investigated in this paper.
These issues are essential in the development of future acquisition functions and the improvement of the existing ones as they encourage an accurate measure of the mutual information such as the \emph{rectified MES} (RMES) acquisition function we develop in this work.
Unlike the evaluation of MES,
we derive a closed-form probability density for the observation conditioned on the max-value and employ stochastic gradient ascent with reparameterization to efficiently optimize RMES.
As a result of a more principled acquisition function, RMES shows a consistent improvement over MES in several synthetic function benchmarks and real-world optimization problems.
\end{abstract}

\section{Introduction}
\emph{Bayesian optimization} (BO) has demonstrated to be highly effective in optimizing an unknown complex objective function (i.e., possibly noisy, non-convex, without a closed-form expression nor derivative) with a finite budget of expensive function  evaluations~\cite{brochu10tut,shahriari15,snoek12}. 
A BO algorithm depends on a choice of acquisition function
(e.g., improvement-based such as probability of improvement   \cite{kushner1964new} and  \emph{expected improvement} (EI) \cite{movckus78}, information-based such as those described below, or \emph{upper confidence bound} (UCB) \cite{srinivas10ucb}) 
as a heuristic to guide its search for the global maximizer. To do this, the BO algorithm utilizes the chosen acquisition function to iteratively select an input query for evaluating the unknown objective function that trades off between observing at or near to a likely maximizer based on a \emph{Gaussian process} (GP) belief of the objective function (exploitation) vs. improving the GP belief (exploration) until the budget is expended.

In this paper, we consider information-based acquisition functions based on the widely celebrated notion of mutual information, which include \emph{entropy search} (ES) \cite{es}, \emph{predictive ES} (PES) \cite{pes}, \emph{output-space PES} (OPES) \cite{hoffman15opes}, \emph{max-value ES} (MES) \cite{wang17mes}, and fast information-theoretic BO \cite{ru2018fast}.
In general, they either
maximize the information gain on the \emph{max-value} (i.e., global maximum of objective function) or its corresponding global \emph{maximizer}.
Though ES and PES perform the latter and hence directly achieve the goal of BO,
they require a series of approximations.
On the other hand, MES and OPES perform the former and  
enjoy the advantage of requiring sampling of only a $1$-dimensional random variable representing the max-value (instead of a multi-dimensional random vector representing the maximizer).
In particular, MES can be expressed in closed form and  
thus optimized easily.
Unfortunately, its BO performance is compromised 
by a number of misconceptions surrounding its design, as discussed below. 
Since there may be subsequent works building on MES (e.g., \cite{knudde2018data,takeno2019multi}), it is imperative that we rectify these pressing misconceptions.

So, our first contribution of this paper is to review the principle of information gain underlying MES, which will shed light on two (perhaps) surprising misconceptions (Section~\ref{sec:mesissue}) and their negative implications on its 
interpretation as a mutual information measure and its BO performance.
We give an intuitive illustration in Section~\ref{sec:rmes} using simple synthetic experiments how they can cause its search to behave suboptimally in trading off between exploitation vs.~exploration.

Our second contribution is the development of 
a \emph{rectified max-value entropy search} (RMES) acquisition function that resolves the misconceptions present in MES and hence provides a more precise measure of mutual information.
In contrast to the straightforward implementation of MES,
the evaluation of RMES may seem challenging at first glance: 
Unlike the noiseless observations assumed by MES that follow the well-known truncated Gaussian distribution when conditioned on the max-value, the true noisy observations obtained from evaluating the objective function do not.
However, by deriving a closed-form expression for the conditional probability density of the noisy observation,
we make the evaluation of RMES possible. Furthermore, the optimization is made efficient by using stochastic gradient ascent with a reparameterization trick \cite{kingma2013auto}.
As a result of a more principled acquisition function, RMES gives a rewarding performance that is improved over the existing MES in several synthetic function benchmarks and real-world optimization problems.

\section{Background}

\subsection{Gaussian Processes in BO}
\label{sec:gpbo}

Consider the problem of sequentially optimizing an unknown objective function $f: \mcl{X} \rightarrow \mbb{R}$ over a bounded input domain $\mcl{X} \subset \mbb{R}^d$.
BO algorithms repeatedly select an input query $\mbf{x} \in \mcl{X}$ for evaluating $f$ to obtain a noisy observed output $y_{\mbf{x}} \triangleq f_{\mbf{x}} + \epsilon$ with $f_{\mbf{x}} \triangleq f(\mbf{x})$, 
i.i.d. Gaussian noise $\epsilon \sim \mcl{N}(0,\sigma_n^2)$, and noise variance $\sigma_n^2$. Since it is costly to evaluate $f$, our goal is to strategically select input queries for finding the maximizer $\mbf{x}_* \triangleq \argmax_{\mbf{x}' \in \mcl{X}} f_{\mbf{x}'}$ as rapidly as possible.
To achieve this, the belief of $f$ is modeled by a GP. Let $\{ f_{\mbf{x}'} \}_{\mbf{x}' \in \mcl{X}}$ denote a GP, that is, every finite subset of $\{ f_{\mbf{x}'} \}_{\mbf{x}' \in \mcl{X}}$ follows a multivariate Gaussian distribution \cite{rasmussen06}. Then, the GP is fully specified by its \emph{prior} mean $\mbb{E}[f_{\mbf{x}'}]$ and covariance $k_{\mbf{x}'\mbf{x}''} \triangleq \text{cov}[f_{\mbf{x}'}, f_{\mbf{x}''}]$ for all $\mbf{x}', \mbf{x}'' \in \mcl{X}$, the latter of
which can be defined, for example, by the widely-used \emph{squared exponential} (SE) kernel
$k_{\mbf{x}'\mbf{x}''} \triangleq \sigma_s^2\ \exp(-0.5(\mbf{x}' - \mbf{x}'')^\top{\Lambda}^{-2}(\mbf{x}' - \mbf{x}''))$
where ${\Lambda} \triangleq \mathrm{diag}[\ell_1, \ldots, \ell_d]$ and $\sigma_s^2$ are its length-scale and signal variance hyperparameters, respectively. For notational simplicity (and w.l.o.g.), the prior mean is assumed to be zero. Given a column vector $\mbf{y}_{\mathcal{D}}\triangleq (y_{\mbf{x}'})^{\top}_{\mbf{x}'\in\mcl{D}}$ of noisy outputs observed from evaluating $f$ at a set $\mcl{D}$ of input queries selected in previous BO iterations, the GP predictive belief of $f$ at any input query $\mbf{x}$ is a Gaussian $f_{\mbf{x}} |\mbf{y}_{\mathcal{D}} \sim \mcl{N}(\mu_{\mbf{x}}, \sigma_{\mbf{x}}^2)$ with the following \emph{posterior} mean $\mu_{\mbf{x}}$ and variance $\sigma_{\mbf{x}}^2$:
\begin{equation}
\begin{array}{l}
\displaystyle \mu_{\mbf{x}} 
	\triangleq 
K_{\mbf{x}\mcl{D}}(K_{\mcl{D}\mcl{D}}+\sigma^2_n I)^{-1}
	\mbf{y}_{\mcl{D}}\\
\displaystyle \sigma_{\mbf{x}}^2 
	\triangleq k_{\mbf{x} \mbf{x}}
	- K_{\mbf{x}\mcl{D}}(K_{\mcl{D}\mcl{D}}+\sigma^2_n I)^{-1}
	K_{\mcl{D}\mbf{x}}
\end{array}
\label{eq:gppost}
\end{equation}
where 
$K_{\mbf{x}\mcl{D}}\triangleq(k_{\mbf{x}\mbf{x}'})_{\mbf{x}'\in \mcl{D}}$, $K_{\mcl{D}\mcl{D}}\triangleq(k_{\mbf{x}'\mbf{x}''})_{\mbf{x}', \mbf{x}''\in \mcl{D}}$, and $K_{\mcl{D}\mbf{x}}\triangleq K^{\top}_{\mbf{x}\mcl{D}}$.
Then, $y_{\mbf{x}} |\mbf{y}_{\mathcal{D}} \sim \mcl{N}(\mu_{\mbf{x}}, \sigma_{+}^2 \triangleq \sigma_{\mbf{x}}^2+\sigma^2_n)$.

\subsection{Max-value information gain}

Mutual information between $2$ random variables has been used to quantify the amount of information gain about one random variable through observing the other. In the context of BO where the observations are the noisy function output $y_{\mbf{x}}$ at the input query $\mbf{x}$, the information gain is about the maximizer (e.g., \emph{entropy search} (ES) \cite{es} and \emph{predictive entropy search} (PES) \cite{pes}), i.e., $\mbf{x}_*$, or the max-value (e.g., \emph{max-value entropy search} (MES) \cite{wang17mes}), denoted as $f_* \triangleq f_{\mbf{x}_*}$. In this paper, we consider the information gain on the max-value.
The information gain on the max-value can be interpreted as the reduction in the uncertainty of the max-value $f_*$ by observing $y_{\mbf{x}}$ where the uncertainty is measured by the entropy, i.e., the mutual information between $f_*$ and $y_{\mbf{x}}$:
\begin{equation}
I(f_*; y_{\mbf{x}} | \mbf{y}_{\mcl{D}})
	= H(p(f_* | \mbf{y}_{\mcl{D}}))
	  - \mbb{E}_{p(y_{\mbf{x}}|\mbf{y}_{\mcl{D}})}[ H(p(f_* | \mbf{y}_{\mcl{D}}, y_{\mbf{x}})) ]
\label{eq:fMI}
\end{equation}
where, given a random variable $z$ following a distribution specified by the probability density $p(z)$, $H(p(z))$ denotes the entropy of  $z$ and $\mbb{E}_{p(z)} f'(z)$ denotes the expectation of a function $f'$ over $z$.
As \eqref{eq:fMI} requires $p(f_*| \mbf{y}_{\mcl{D}}, y_{\mbf{x}})$ which is computationally expensive to evaluate for different values of $y_{\mbf{x}}$, the symmetry property of the mutual information is often exploited to express the acquisition function as:
\begin{equation}
I(f_*; y_{\mbf{x}} | \mbf{y}_{\mcl{D}})
	= H(p(y_{\mbf{x}} | \mbf{y}_{\mcl{D}}))
	  - \mbb{E}_{p(f_*|\mbf{y}_{\mcl{D}})}[ H(p(y_{\mbf{x}} | \mbf{y}_{\mcl{D}}, f_*)) ]\ .
\label{eq:yMI}
\end{equation}
To evaluate the acquisition function in \eqref{eq:yMI}, it requires the evaluation of 
$p(y_{\mbf{x}} | \mbf{y}_{\mcl{D}}, f_*)$.
This probability is difficult to evaluate as it requires imposing the condition that $f_{\mbf{x}'} \le f_*\ \forall \mbf{x}' \in \mcl{X}$, so MES only imposes the condition of the max-value at the input query:\footnote{It is noted that OPES uses a more relaxed assumption than MES, but it requires a difficult approximation.}
\begin{equation}
p(y_{\mbf{x}} | \mbf{y}_{\mcl{D}}, f_*) \approx \int p(y_{\mbf{x}} | f_{\mbf{x}}) p(f_{\mbf{x}} | \mbf{y}_{\mcl{D}}) \mbb{I}_{f_{\mbf{x}} \le f_*}\ \text{d}f_{\mbf{x}}
\label{eq:approxpy}
\end{equation}
where $\mbb{I}_{f_{\mbf{x}} \le f_*}$ is the indicator function such that it is $1$ if $f_{\mbf{x}} \le f_*$ and $0$ otherwise. We adopt this assumption throughout the paper.

Furthermore, to make \eqref{eq:approxpy} a truncated Gaussian density whose entropy has a closed-form expression, MES replaces $y_{\mbf{x}}$ with $f_{\mbf{x}}$ on the right hand side of \eqref{eq:approxpy}, which results in $\alpha_{\text{MES}}(\mbf{x}, \mbf{y}_{\mcl{D}}) \triangleq I(f_*; f_{\mbf{x}} | \mbf{y}_{\mcl{D}}) = $
\begin{equation}
H(p(f_{\mbf{x}} | \mbf{y}_{\mcl{D}}))
	  - \mbb{E}_{p(f_*|\mbf{y}_{\mcl{D}})}[ H(p(f_{\mbf{x}} | \mbf{y}_{\mcl{D}}, f_*)) ]\ .
\label{eq:mes}
\end{equation}
This formulation of MES has been applied in quite a few works, e.g.,  \cite{knudde2018data,takeno2019multi}. In \eqref{eq:mes}, $f_{\mbf{x}}|\mbf{y}_{\mcl{D}}, f_*$ follows a truncated Gaussian distribution whose entropy has a closed-form expression. Hence, the MES expression can be reduced to \cite{wang17mes}:
\begin{equation}
\frac{1}{|\mcl{F}|} \sum_{f_* \in \mcl{F}} \left[ 
	\frac{h_{f_*}(\mbf{x}) \psi(h_{f_*}(\mbf{x}))}{2 \Psi(h_{f_*}(\mbf{x}))} - \log \Psi(h_{f_*}(\mbf{x}))
\right]
\label{eq:mesdetail}
\end{equation}
where $\mcl{F}$ is a finite set of samples of $f_*$ drawn from $p(f_*|\mbf{y}_{\mcl{D}})$, $\displaystyle h_{f_*}(\mbf{x}) \triangleq \frac{f_* - \mu_{\mbf{x}}}{\sigma_{\mbf{x}}}$, $\psi(h_{f_*}(\mbf{x})) \triangleq \mcl{N}(h_{f_*}(\mbf{x}); 0,1)$ denotes the probability density function at $h_{f_*}(\mbf{x})$ of the standard Gaussian distribution, and $\Psi(h_{f_*}(\mbf{x}))$ denotes the cumulative density function value at $h_{f_*}(\mbf{x})$ of the standard Gaussian distribution. This set $\mcl{F}$ is obtained by either optimizing sampled functions from the GP posterior \cite{pes,wang17mes} or approximating with a Gumbel distribution \cite{wang17mes}.

\section{Misconceptions in MES}
\label{sec:mesissue}

This section investigates two main issues with the existing MES \cite{wang17mes}, which paves the way to an improved variant of MES in the next section. 

\subsection{Noiseless Observations}
\label{sec:noiseless}

Since BO observations are often the noisy function output $y_{\mbf{x}}$ due to uncontrolled factors such as random noise in environmental sensing, stochastic optimization, and random mini-batches of data in machine learning, replacing $y_{\mbf{x}}$ with $f_{\mbf{x}}$ in \eqref{eq:mes} fundamentally changes the principle behind the information gain on the max-value.
In other words, MES measures the amount of information gain about $f_*$ through observing $f_{\mbf{x}}$ which is often not observed in practice.
In fact, the noisy observation $y_{\mbf{x}}$ contains less information about the latent function than its noiseless counterpart $f_{\mbf{x}}$.
Thus, replacing $y_{\mbf{x}}$ with $f_{\mbf{x}}$ potentially overestimates the amount of information gain as shown in Fig.~\ref{fig:comexploit} in Section~\ref{sec:rmes}.

\begin{figure}
\centering
\begin{tabular}{@{}c@{}c@{}}
\includegraphics[height=0.2\textwidth]{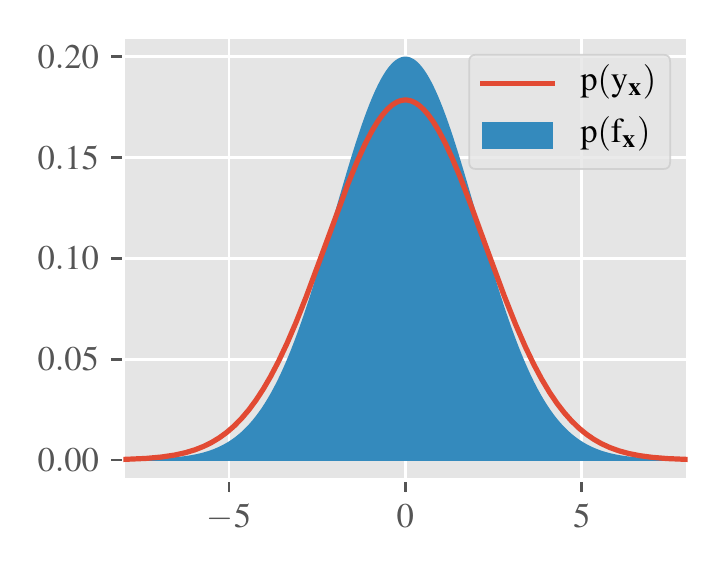}
&
\includegraphics[height=0.2\textwidth]{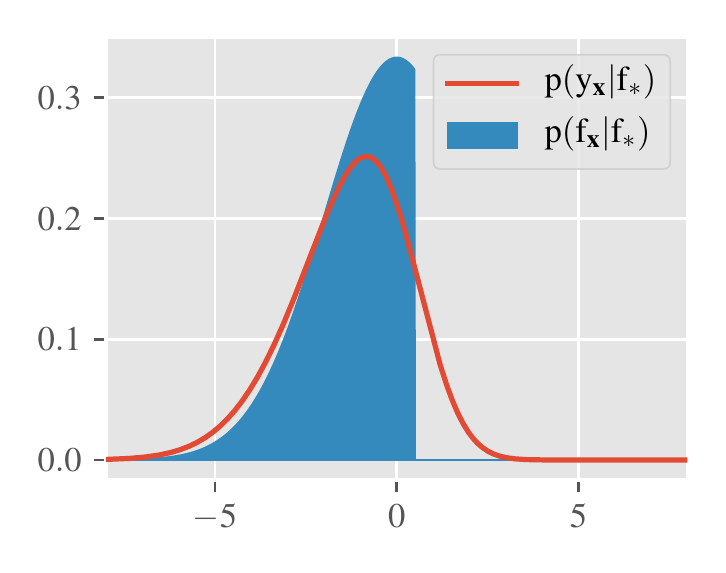}
\\
(a) $p(f_{\mbf{x}})$ vs. $p(y_{\mbf{x}})$.
&
(b) $p(f_{\mbf{x}} | f_*)$ vs. $p(y_{\mbf{x}} | f_*)$.
\end{tabular}
\caption{A comparison between the distribution of the noiseless function function output $f_{\mbf{x}}$ and that of the noisy function output $y_{\mbf{x}}$ where $f_{\mbf{x}} \sim \mcl{N}(0,4)$, $y_{\mbf{x}} = f_{\mbf{x}} + \epsilon$, $\epsilon \sim \mcl{N}(0,1)$, and $f_* = 0.5$.}
\label{fig:tnvsntn}
\end{figure}

Fig.~\ref{fig:tnvsntn} illustrates the difference between the distribution of the noisy function output $y_{\mbf{x}}$ (i.e., the observation of BO) and that of the noiseless function output $f_{\mbf{x}}$. It is noted that conditioned on the max-value $f_*$, the distribution of the noiseless function output $f_{\mbf{x}}|f_*$ is a truncated Gaussian distribution having a closed-form expression for its entropy. 
On the contrary, it is challenging to evaluate the probability density of the noisy function output conditioned on the max-value, i.e., $y_{\mbf{x}}|f_*$ in Fig.~\ref{fig:tnvsntn}b, not to mention its entropy.
We will resolve this issue in Section~\ref{sec:rmes}.

\subsection{Discrepancy in Evaluation}
\label{sec:discrepancy}

The mutual information can be interpreted as the mutual dependence between two random variables via the \emph{Kullback-Leibler} (KL) divergence.
In \eqref{eq:mes}, these random variables are $f_{\mbf{x}}$ and $f_*$, and the mutual information can be expressed as $I(f_*; f_{\mbf{x}} | \mbf{y}_{\mcl{D}}) = \text{D}_{\text{KL}}\left[p(f_*,f_{\mbf{x}} | \mbf{y}_{\mcl{D}})\ \Vert\ p(f_*|\mbf{y}_{\mcl{D}})p(f_{\mbf{x}} | \mbf{y}_{\mcl{D}})\right]$ which denotes the KL divergence between $p(f_*,f_{\mbf{x}} | \mbf{y}_{\mcl{D}})$ and $p(f_*|\mbf{y}_{\mcl{D}})p(f_{\mbf{x}} | \mbf{y}_{\mcl{D}})$.
If the KL divergence between $p(f_*,f_{\mbf{x}} | \mbf{y}_{\mcl{D}})$ and the fully factorized distribution $p(f_*|\mbf{y}_{\mcl{D}})p(f_{\mbf{x}} | \mbf{y}_{\mcl{D}})$ is large, $f_*$ and $f_{\mbf{x}}$ are dependent on each other.
Hence, the distributions of $f_*$ and $f_{\mbf{x}}$ should be defined consistently throughout the evaluation of the mutual information.
On the contrary, if there exist discrepancies in the evaluation of the mutual information such that $p(f_{\mbf{x}}|\mbf{y}_{\mcl{D}})$ or $p(f_*|\mbf{y}_{\mcl{D}})$ is not uniquely evaluated, the mutual information is no longer properly defined.
Therefore, for MES to have an interpretation as a measure of the mutual information, $p(f_*|\mbf{y}_{\mcl{D}})$ and $p(f_{\mbf{x}}|\mbf{y}_{\mcl{D}})$ should be consistently defined throughout the MES evaluation. For example, in \eqref{eq:mes}, $p(f_{\mbf{x}}|\mbf{y}_{\mcl{D}})$ should have the same probability density under assumptions in $H(p(f_{\mbf{x}}|\mbf{y}_{\mcl{D}}))$ and those in $\mbb{E}_{p(f_*|\mbf{y}_{\mcl{D}})}[ H(p(f_{\mbf{x}} | \mbf{y}_{\mcl{D}}, f_*)) ]$.

Let us consider the conditional probability density $p(f_{\mbf{x}} | \mbf{y}_{\mcl{D}})$ under assumptions imposed when evaluating $H(p(f_{\mbf{x}}|\mbf{y}_{\mcl{D}}))$ and $\mbb{E}_{p(f_*|\mbf{y}_{\mcl{D}})}[ H(p(f_{\mbf{x}} | \mbf{y}_{\mcl{D}}, f_*)) ]$ of MES in order to identify the discrepancy in the evaluation of these two terms.

\begin{itemize}
\item Evaluating $\displaystyle H(p(f_{\mbf{x}}|\mbf{y}_{\mcl{D}}))$: In MES, there is no approximation when evaluating $\displaystyle H(p(f_{\mbf{x}}|\mbf{y}_{\mcl{D}}))$ since $p(f_{\mbf{x}}|\mbf{y}_{\mcl{D}})$ is the density of a Gaussian distribution with a closed-form expression for its entropy, i.e.,
\begin{equation}
f_{\mbf{x}}|\mbf{y}_{\mcl{D}} \sim \mcl{N}(\mu_{\mbf{x}}, \sigma_{\mbf{x}}^2)\ .
\label{eq:fxdist1}
\end{equation}
\item Evaluating $\mbb{E}_{p(f_*|\mbf{y}_{\mcl{D}})}[ H(p(f_{\mbf{x}} | \mbf{y}_{\mcl{D}}, f_*)) ]$: As the expectation over $f_*|\mbf{y}_{\mcl{D}}$ is intractable, MES approximates the expectation with an average over a finite set $\mcl{F}$ of max-value samples, i.e., $\displaystyle \mbb{E}_{p(f_*|\mbf{y}_{\mcl{D}})} [H(p(f_{\mbf{x}} | \mbf{y}_{\mcl{D}}, f_*)) ] \approx \frac{1}{|\mcl{F}|} \sum_{f_* \in \mcl{F}} H(p(f_{\mbf{x}} | \mbf{y}_{\mcl{D}}, f_*))$. This set $\mcl{F}$ is obtained in the same approach in \eqref{eq:mesdetail}.
Under this assumption, it follows that
\begin{equation}
f_{\mbf{x}}|\mbf{y}_{\mcl{D}} \sim \frac{1}{|\mcl{F}|} \sum_{f_* \in \mcl{F}} \mcl{N}_{f_*}(\mu_{\mbf{x}}, \sigma_{\mbf{x}}^2)
\label{eq:fxdist2}
\end{equation}
where $\mcl{N}_{f_*}(\mu_{\mbf{x}}, \sigma_{\mbf{x}}^2)$ denotes a distribution obtained from restricting a Gaussian distribution $\mcl{N}(\mu_{\mbf{x}}, \sigma_{\mbf{x}}^2)$ to the interval of $(-\infty, f_*]$, i.e., an upper truncated Gaussian distribution.
\end{itemize}

\begin{figure}
\centering
\includegraphics[width=0.3\textwidth]{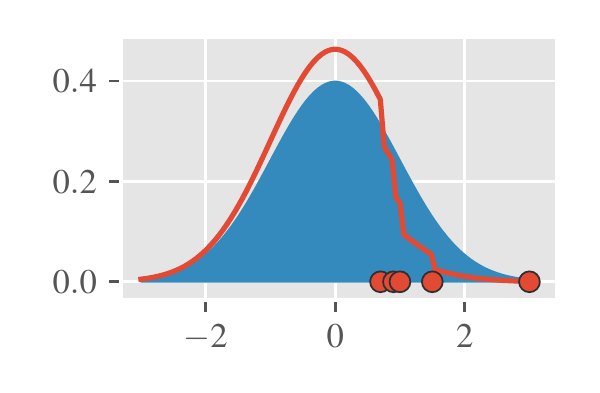}
\caption{Difference between $\mcl{N}(0,1)$ (shade blue area) and $\sum_{u} \mcl{N}_{u}(0,1) / 4$ (red line) for $u \in \{0.7, 0.9, 1, 1.5, 3\}$ (red circles).}
\label{fig:diff}
\end{figure}

At first glance, the use of the Gaussian distribution in \eqref{eq:fxdist1} is convenient and fairly straightforward for the closed-form expression of its entropy.
However, putting the probability densities inferred from the evaluation of the two terms (i.e., \eqref{eq:fxdist1} and \eqref{eq:fxdist2}) together, it becomes obvious that 
there exists a discrepancy in the evaluation of MES as $p(f_{\mbf{x}} | \mbf{y}_{\mcl{D}})$ is different under assumptions imposed on $H(p(f_{\mbf{x}}|\mbf{y}_{\mcl{D}}))$ and those imposed on $\mbb{E}_{p(f_*|\mbf{y}_{\mcl{D}})}[ H(p(f_{\mbf{x}} | \mbf{y}_{\mcl{D}}, f_*)) ]$.
This is because \eqref{eq:fxdist1} is the result of marginalizing out $f_*$ for all of its possible values in $\mbb{R}$, while \eqref{eq:fxdist2} is the result of marginalizing out $f_*$ over a finite set $\mcl{F} \subset \mbb{R}$. 
The discrepancy between a Gaussian distribution and the average over truncated Gaussian distributions is illustrated in Fig.~\ref{fig:diff}. 
This difference violates the definition of a random variable as $f_{\mbf{x}} | \mbf{y}_{\mcl{D}}$ does not have a unique distribution throughout the evaluation of MES. 
As a result, MES cannot be interpreted as the mutual dependence (mutual information) between $2$ random variables since there is not a consistent description for the distribution of $f_{\mbf{x}} | \mbf{y}_{\mcl{D}}$ in its evaluation. 
An undesired consequence is that MES might over-explore as shown in Fig.~\ref{fig:comexplore} in Section~\ref{sec:rmes}.
It is noted that a similar issue also exists in PES \cite{pes}.

\section{Rectified Max-Value Entropy Search}
\label{sec:rmes}

In this section, we propose a \emph{rectified max-value entropy search} (RMES) measuring the information gain $I(f_*;y_{\mbf{x}} | \mbf{y}_{\mcl{D}})$ without the above misconceptions, i.e.,
$\alpha_{\text{RMES}}(\mbf{x}, \mbf{y}_{\mcl{D}}) \triangleq I(f_*;y_{\mbf{x}} | \mbf{y}_{\mcl{D}}) = H(p(y_{\mbf{x}} | \mbf{y}_{\mcl{D}})) - \mbb{E}_{p(f_*|\mbf{y}_{\mcl{D}})}[
	H(p(y_{\mbf{x}} | \mbf{y}_{\mcl{D}}, f_*))
]$. 
The expectation over $f_*|\mbf{y}_{\mcl{D}}$ is approximated with a finite set $\mcl{F}$ of samples obtained by optimizing sampled functions from the GP posterior \cite{pes,wang17mes}.

In MES, $f_{\mbf{x}}| \mbf{y}_{\mcl{D}}, f_*$ follows a truncated Gaussian distribution whose entropy has a closed-form expression.
On the contrary, in RMES, $y_{\mbf{x}} | \mbf{y}_{\mcl{D}}, f_*$ is the sum of a Gaussian random variable $\epsilon$ and a truncated Gaussian random variable $f_{\mbf{x}} | \mbf{y}_{\mcl{D}}, f_*$ ($p(y_{\mbf{x}}|f_*))$ in Fig.~\ref{fig:tnvsntn}b).
Optimizing RMES which involves the entropy of 
$y_{\mbf{x}} | \mbf{y}_{\mcl{D}}, f_*$ is challenging as no analytical expression of the entropy is available.
To resolve this challenge, we first derive a closed-form expression for the probability density of 
$y_{\mbf{x}} | \mbf{y}_{\mcl{D}}, f_*$ in the following theorem.

\begin{theorem} \label{theorem:ntn}
The probability density function of $y_{\mbf{x}} | \mbf{y}_{\mcl{D}}, f_*$ is expressed as:
\begin{equation}
p(y_{\mbf{x}} | \mbf{y}_{\mcl{D}}, f_*) = \mcl{N}(y_{\mbf{x}}; \mu_{\mbf{x}}, \sigma_{+}^2) 
	\frac{
	\Psi(g_{f_*}(y_{\mbf{x}}))
	}
	{
	\Psi(h_{f_*}(\mbf{x}))
	}
\label{eq:py}
\end{equation}
where $\displaystyle g_{f_*}(y_{\mbf{x}}) \triangleq \left( \sigma_+^2 f_*  - \sigma_n^2 \mu_{\mbf{x}} - \sigma_{\mbf{x}}^2 y_{\mbf{x}}\right) / \left(\sigma_{\mbf{x}} \sigma_n  \sigma_+\right)$; 
$\sigma_{+}^2 \triangleq \sigma_{\mbf{x}}^2 + \sigma_n^2$;
$ h_{f_*}(\mbf{x})$ and $\Psi$ are specified in \eqref{eq:mesdetail}.
\end{theorem}

The derivation of \eqref{eq:py} is in Appendix~\ref{app:pdfntn}.
Although we cannot evaluate the entropy of $y_{\mbf{x}}|\mbf{y}_{\mcl{D}}, f_*$ analytically, we can optimize it using the stochastic gradient ascent.
Given the closed-form expression in \eqref{eq:py}, a straightforward solution is to express $H(p(y_{\mbf{x}} | \mbf{y}_{\mcl{D}}, f_*))$ as $\mbb{E}_{ p(y_{\mbf{x}} | \mbf{y}_{\mcl{D}}, f_*)}[- \log  p(y_{\mbf{x}} | \mbf{y}_{\mcl{D}}, f_*)]$. Then, one can sample batches of $ y_{\mbf{x}} | \mbf{y}_{\mcl{D}}, f_*$ by first sampling $f_{\mbf{x}} | \mbf{y}_{\mcl{D}}, f_*$ from a truncated Gaussian distribution, then, sampling the Gaussian noise $\epsilon$, and summing them up. 
Stochastic gradient optimization has been made feasible for an expectation over a random sample following a truncated Gaussian distribution by a clever trick recently, namely, implicit reparameterization gradients \cite{figurnov2018implicit}. 
However, as samples of $y_{\mbf{x}} | \mbf{y}_{\mcl{D}}, f_*$ depend on $f_*$, we need to draw different samples of $y_{\mbf{x}} | \mbf{y}_{\mcl{D}}, f_*$  for different values of $f_* \in \mcl{F}$, which is inefficient.

Fortunately, we can design a more efficient approach to stochastically optimize $H(p(y_{\mbf{x}} | \mbf{y}_{\mcl{D}}, f_*))$ where samples are drawn independently from $f_*$ and the optimization only requires the simple reparameterization trick: the Gaussian standardization. By factoring $p(y_{\mbf{x}} | \mbf{y}_{\mcl{D}}, f_*)$ into $w_{f_*}(y_{\mbf{x}}) \triangleq \Psi(g_{f_*}(y_{\mbf{x}})) / \Psi(h_{f_*}(\mbf{x}))$ dependent on $f_*$ and $p(y_{\mbf{x}}|\mbf{y}_{\mcl{D}}) = \mcl{N}(y_{\mbf{x}}; \mu_{\mbf{x}}, \sigma_{+}^2)$  independent from $f_*$. 
We can express $H(p(y_{\mbf{x}} | \mbf{y}_{\mcl{D}}, f_*))$ in an importance sampling approach with the weight $w_{f_*}(y_{\mbf{x}})$ as 
\begin{equation*}
H(p(y_{\mbf{x}} | \mbf{y}_{\mcl{D}}, f_*)) = \mbb{E}_{p(y_{\mbf{x}}|\mbf{y}_{\mcl{D}})}\left[- w_{f_*}(y_{\mbf{x}}) \log p(y_{\mbf{x}}| \mbf{y}_{\mcl{D}}, f_*) \right]
\end{equation*}
where $p(y_{\mbf{x}}| \mbf{y}_{\mcl{D}}, f_*)$ is computed in \eqref{eq:py}.
The expectation over the Gaussian distribution $p(y_{\mbf{x}}|\mbf{y}_{\mcl{D}})$ which is independent from $f_*$ can be reparameterized using the Gaussian standardization in \cite{kingma2013auto}:
\begin{equation*}
H(p(y_{\mbf{x}} | \mbf{y}_{\mcl{D}}, f_*)) = \mbb{E}_{p(\nu)}\left[ -w_{f_*}(t(\nu)) \log p(t(\nu)| \mbf{y}_{\mcl{D}}, f_*) \right]
\end{equation*}
where $\nu \sim \mcl{N}(0,1)$, $t(\nu) \triangleq \nu \sigma_{+} + \mu_{\mbf{x}}$, and $p(t(\nu)| \mbf{y}_{\mcl{D}}, f_*) \triangleq p(y_{\mbf{x}} = t(\nu)| \mbf{y}_{\mcl{D}}, f_*)$. The expectation over $\nu$ does not depend on $\mbf{x}$ nor $f_*$, so samples of $\nu$ can be shared between different samples of $f_*$.
Furthermore, the gradient of the acquisition function is not propagated through this sampling procedure.
Hence, the term $\displaystyle \mbb{E}_{p(f_*|\mbf{y}_{\mcl{D}})}\left[ H(p(y_{\mbf{x}} | \mbf{y}_{\mcl{D}}, f_*)) \right] \approx \frac{1}{|\mcl{F}|} \sum_{f_* \in \mcl{F}} H(p(y_{\mbf{x}} | \mbf{y}_{\mcl{D}}, f_*))$ in RMES can be expressed as:
\begin{equation}
\mbb{E}_{p(\nu)}\left[- \frac{1}{|\mcl{F}|} \sum_{f_* \in \mcl{F}} w_{f_*}(t(\nu)) \log p(t(\nu)| \mbf{y}_{\mcl{D}}, f_*) \right]\ .
\label{eq:rmescondterm}
\end{equation}

Regarding the term $H(p(y_{\mbf{x}}|\mbf{y}_{\mcl{D}}))$ in RMES, to avoid the discrepancy in the evaluation of MES (Section~\ref{sec:discrepancy}), we apply the same restriction of $f_*$ to the finite set $\mcl{F}$ onto $H(p(y_{\mbf{x}}|\mbf{y}_{\mcl{D}}))$ to obtain
\begin{align*} 
H(p(y_{\mbf{x}} | \mbf{y}_{\mcl{D}})) &= \mbb{E}_{p(\nu)}\Bigg[ -
	\frac{1}{|\mcl{F}|}
		\sum_{f_* \in \mcl{F}}
		w_{f_*}(t(\nu))\\
	&\quad \times \log \left(
	\frac{1}{|\mcl{F}|}
	\sum_{f'_* \in \mcl{F}}
	p(t(\nu)| \mbf{y}_{\mcl{D}}, f'_*)  
\right)
\Bigg]\ .
\end{align*}
where the Gaussian standardization is used to reparameterize $y_{\mbf{x}}$.
Hence, from the above equation and \eqref{eq:rmescondterm}, we have $\alpha_{\text{RMES}}(\mbf{x}, \mbf{y}_{\mcl{D}}) \triangleq I(f_*; y_{\mbf{x}} | \mbf{y}_{\mcl{D}}) = $
\begin{equation}
\mbb{E}_{p(\nu)} 
	\Bigg[ 
		\frac{1}{|\mcl{F}|} \sum_{f_* \in \mcl{F}} w_{f_*}(t(\nu))
		\log \frac{|\mcl{F}| p(t(\nu)| \mbf{y}_{\mcl{D}}, f_*)}
				{
			\sum_{f'_* \in \mcl{F}}
			p(t(\nu)| \mbf{y}_{\mcl{D}}, f'_*)				
				}
	\Bigg]
\label{eq:rmesd}
\end{equation}
which can be optimized using a stochastic optimization algorithm such as Adam \cite{kingma15adam}. It is noted that in \eqref{eq:rmesd}, samples of $\nu$ are shared for different samples of the max-value $f_*$ and both terms in \eqref{eq:yMI}. Hence, the number of samples of $\nu$ can be reduced in comparison with the approach of directly sampling $y_{\mbf{x}} | \mbf{y}_{\mcl{D}}, f_*$ mentioned above.

\begin{figure}[h!]
\centering
\includegraphics[width=0.38\textwidth]{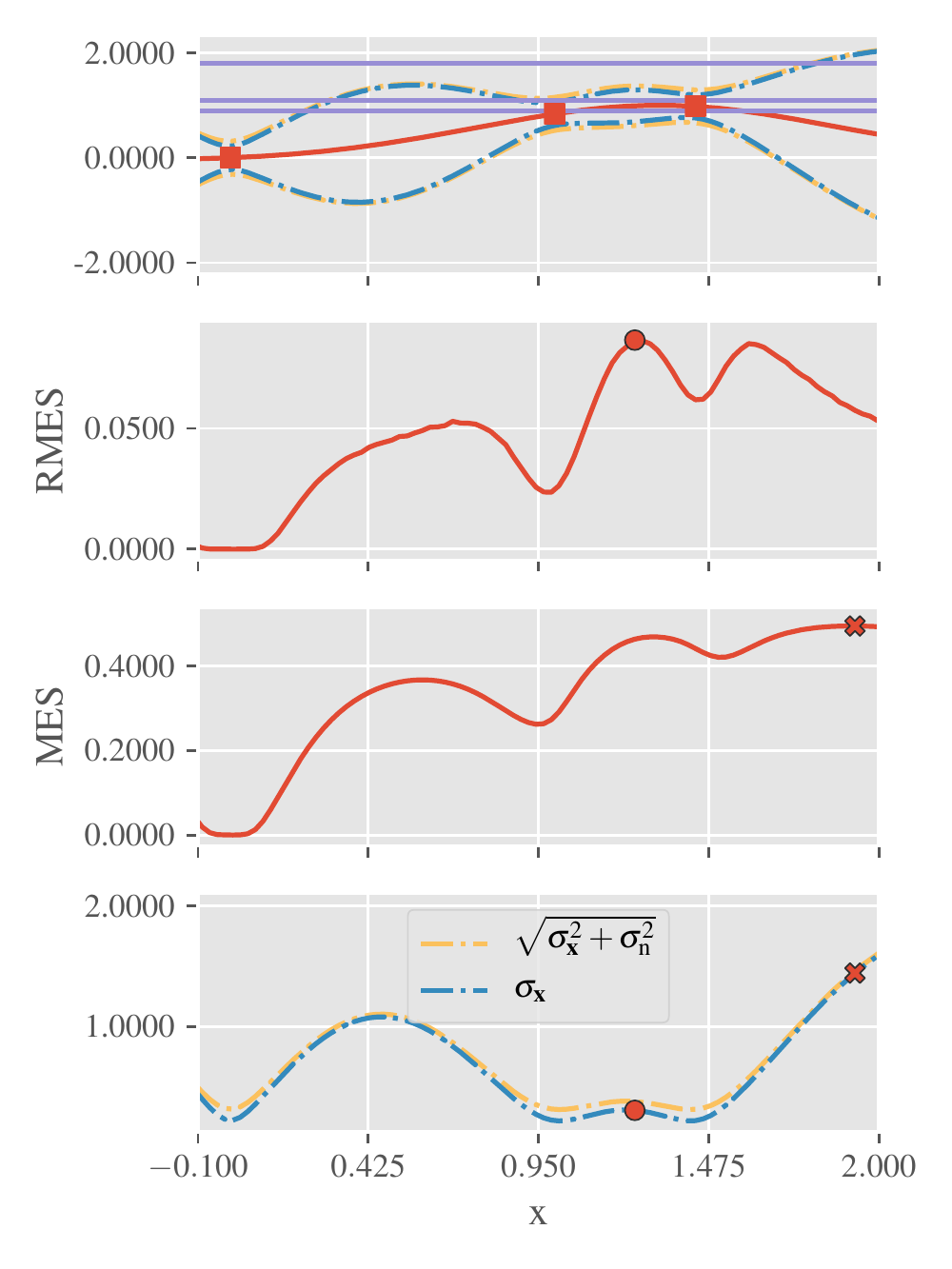}
\caption{Example where MES tends towards an exploration strategy. The plots share the x-axis.
The top plot shows the data samples as red squares; the GP posterior mean as a red line; the GP posterior standard deviations of $y_{\mbf{x}}$ and $f_{\mbf{x}}$ as dashed yellow lines and dashed blue lines, respectively; and the max-value samples as purple lines. The middle two plots show the values of RMES and MES acquisition functions with the maximum values of RMES and MES (the input queries) as a circle and a cross, respectively. The inputs that maximize RMES and MES are also shown in the bottom plot as a circle and a cross, respectively.}
\label{fig:comexplore}
\end{figure}

\begin{figure}[h!]
\centering
\includegraphics[width=0.38\textwidth]{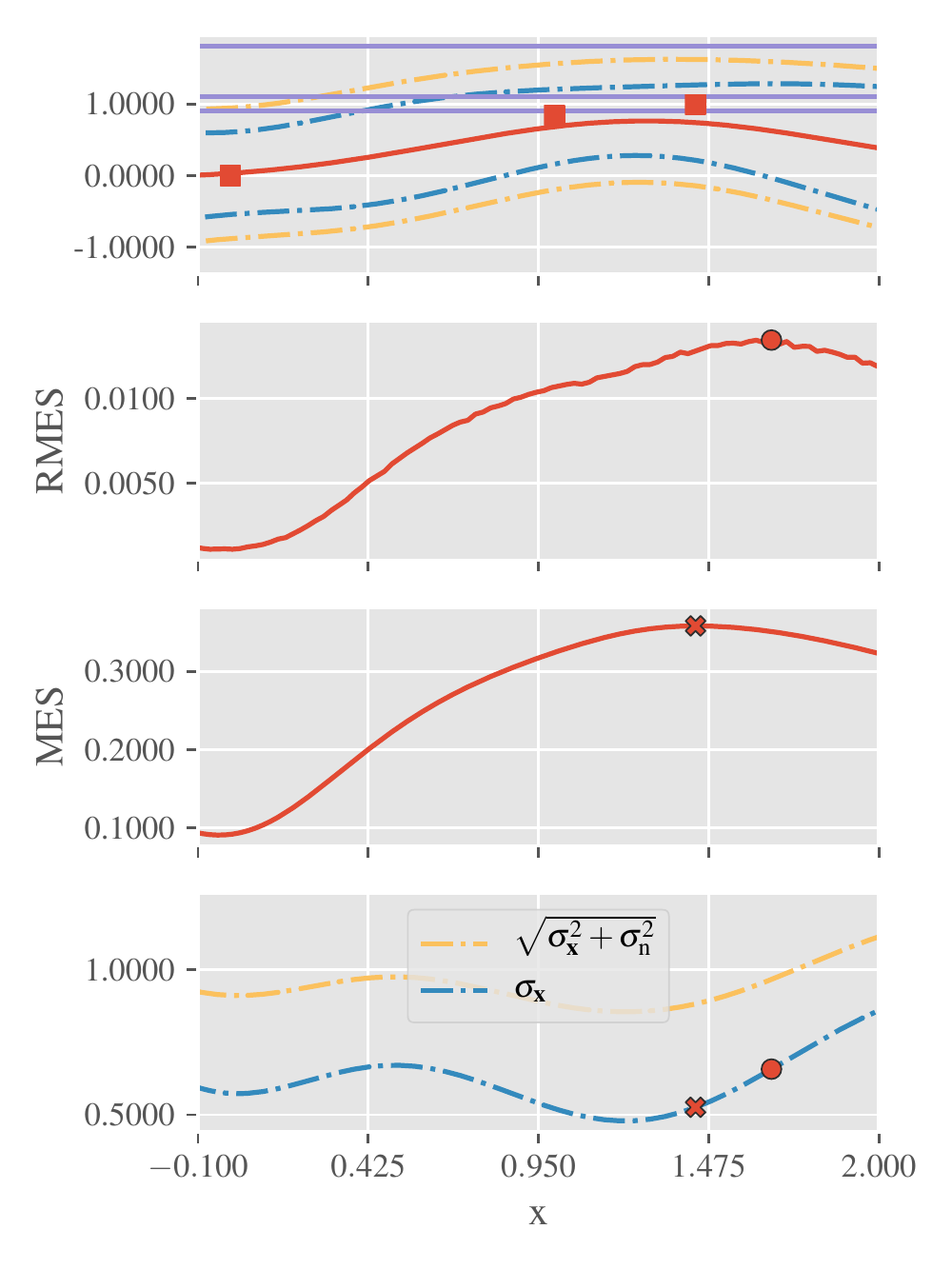}
\caption{Example where MES tends towards an exploitation strategy. The notations are adopted from Fig.~\ref{fig:comexplore}.}
\label{fig:comexploit}
\end{figure}

As mentioned in Section~\ref{sec:mesissue}, the misconceptions in MES cause an imbalance in the exploration-exploitation trade-off. 
To observe the effects of correcting these misconceptions in RMES, let us investigate two simple examples where RMES and MES select different input queries such that  MES over-explores and over-exploits in Fig.~\ref{fig:comexplore} and Fig.~\ref{fig:comexploit}, respectively.

Fig.~\ref{fig:comexplore} illustrates an example where MES tends towards an exploration strategy. In this figure, the noise variance is small to minimize the effect of the noise in the observation, i.e., the noiseless observation issue in Section~\ref{sec:noiseless}. Hence, the dashed blue and dashed yellow lines in Fig.~\ref{fig:comexplore} are mostly overlapped since the standard deviation values of $y_{\mbf{x}}$ and $f_{\mbf{x}}$ are almost the same. We observe that MES selects an input query (the cross) whose function output has a higher posterior variance and a smaller posterior mean than RMES. In other words, MES tends towards exploration. However, MES explores at a cost of not exploiting an uncertain input with a high posterior mean which is selected by RMES (the circle). This could be because the discrepancy in the evaluation issue in Section~\ref{sec:discrepancy} causes the uncertainty in the GP posterior (i.e., the term $H(p(y_{\mbf{x}}|\mbf{y}_{\mcl{D}}))$) to have a high influence on MES. As a result, MES selects an input query far away from the data samples in this example. Therefore, it is noted that a small noise variance does not guarantee a good selection strategy of MES due to the discrepancy in the evaluation. An undesired consequence of this over-exploration is that MES does not query inputs close to the maximizer compared with RMES as shown in the experiments in Section~\ref{sec:exp}.

Fig.~\ref{fig:comexploit} illustrates an example where MES tends towards an exploitation strategy. In this example, the length-scale is set such that function values in Fig.~\ref{fig:comexploit} are more correlated with one another in comparison with those in Fig.~\ref{fig:comexplore}. The noise variance is set to a larger value such that the variance of $y_{\mbf{x}}$ is significantly larger than that of $f_{\mbf{x}}$. Hence, there is a gap between the dashed blue and dashed yellow lines in the top and bottom plots of Fig.~\ref{fig:comexploit}.
MES selects an input query (the cross) whose function value has a smaller posterior variance and a larger posterior mean than RMES. In other words, MES tends towards exploitation. However, MES exploits at an input query where the uncertainty of the noise overwhelms the uncertainty of the unknown objective function since the yellow line is much larger than the blue line at the cross in the bottom plot. This could be because MES overestimates the amount of information gain by not taking into account the noise, i.e., replacing the noisy observation $y_{\mbf{x}}$ with a noiseless function output $f_{\mbf{x}}$ in the acquisition function (Section~\ref{sec:noiseless}). On the other hand, RMES selects an input query (the circle) that balances between the noise in $y_{\mbf{x}}$ and information about $f_*$. Over-exploitation could prevent MES from leaving a suboptimal maximizer, which causes its poor performance in the experiments in Section~\ref{sec:exp}.

\section{Experiments}
\label{sec:exp}

In this section, we empirically evaluate the performance of our RMES and existing acquisition functions such as EI \cite{movckus78}, UCB \cite{srinivas10ucb}, PES \cite{pes}, and MES \cite{wang17mes}. Similar to the MES work \cite{wang17mes}, we use $2$ evaluation criteria: \emph{simple regret} (SR) and \emph{inference regret} (IR). The simple regret measures the regret of the best input query so far, i.e., $f_* - \max_{\mbf{x}' \in \mcl{D}} f_{\mbf{x}'}$. The inference regret measures the regret of the inferred maximizer which is often defined as the maximizer of the GP posterior mean function \cite{es,pes,wang17mes}. In other words, the inference regret is defined as $f_* - f_{\argmax_{\mbf{x}' \in \mcl{X}} \mu_{\mbf{x}'}}$ where $\mu_{\mbf{x}}$ is defined in \eqref{eq:gppost}.

The experiments include (1) synthetic function benchmarks such as a function sample drawn a GP, the Branin-Hoo function, the $2$-dimensional Michaelwicz function, and the eggholder function which is a difficult function to optimize due to its many local maxima; and (2) real-world optimization problems. As an environmental sensing problem, we use the pH field of Broom's Barn farm \cite{webster07} which is spatially distributed over a $1200$m by $680$m region discretized into $31 \times 18$ grid of sampling locations.
To generate a continuous objective function from the dataset, we fit a GP model to the dataset and use its mean function as the objective function which is unknown to all BO algorithms.
For tuning hyperparameters of a machine learning model, we compare the performance of different acquisition functions to tune the hyperparameters of a \emph{support vector machine} (SVM) to fit the Wisconsin breast cancer dataset from the UCI repository \cite{dua2019}.
There are two hyperparameters of the SVM: the penalty parameter of the error term in the range $[0.5,2]$ and the natural logarithm of the kernel coefficient for the radial basis function kernel of the SVM in the range $[-5,-3]$.
Given the SVM's hyperparameters, the unknown objective function is the $100$-fold cross-validation accuracy and the noisy observation is the $20$-fold cross-validation accuracy. The latter is provided to BO as observations at input queries.

To account for the randomness in the observations, synthetic experiments are repeated $15$ times and the real-word experiments are repeated $10$ times with random initialization of $2$ training data samples. The logarithm to the base $10$ of the performance measure average is reported.
The objective functions are each modeled as a sample of a GP whose kernel hyperparameters are learned using maximum likelihood estimation \cite{rasmussen06}. The zero mean and the SE kernel are used. The objective functions are shifted to have a zero mean.

For MES and RMES, there are $5$ samples of the max-value at each BO iteration. For PES, there are $5$ samples of the maximizer at each BO iteration.
The max-value and maximizer samples are drawn by optimizing function samples from the GP posterior \cite{pes,wang17mes}. The approximated sampling of the max-value via a Gumbel distribution \cite{wang17mes} is not used since we do not want the approximation quality to tamper with the performance of the acquisition functions.

\subsection{Synthetic Function Benchmarks}

\begin{figure}[h!]
\centering
\begin{tabular}{@{}c@{}c@{}}
	\includegraphics[height=0.215\textwidth]{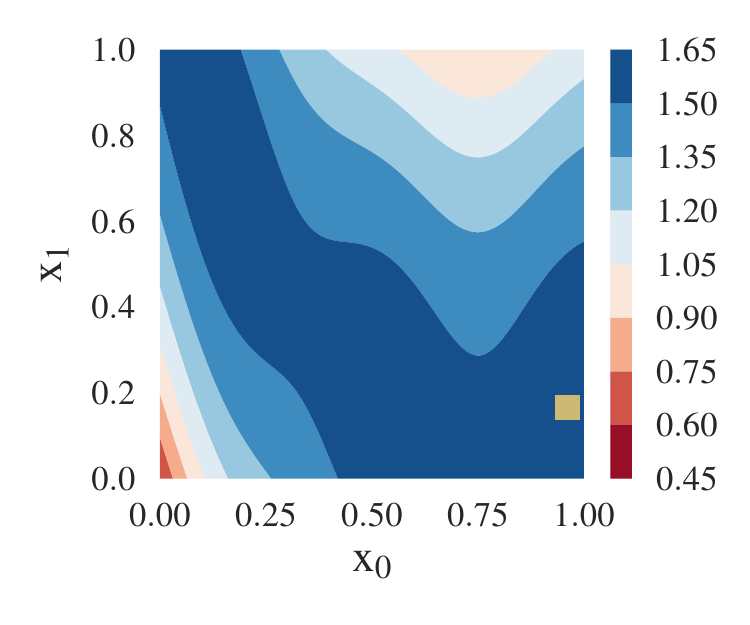}
	&
	\includegraphics[height=0.215\textwidth]{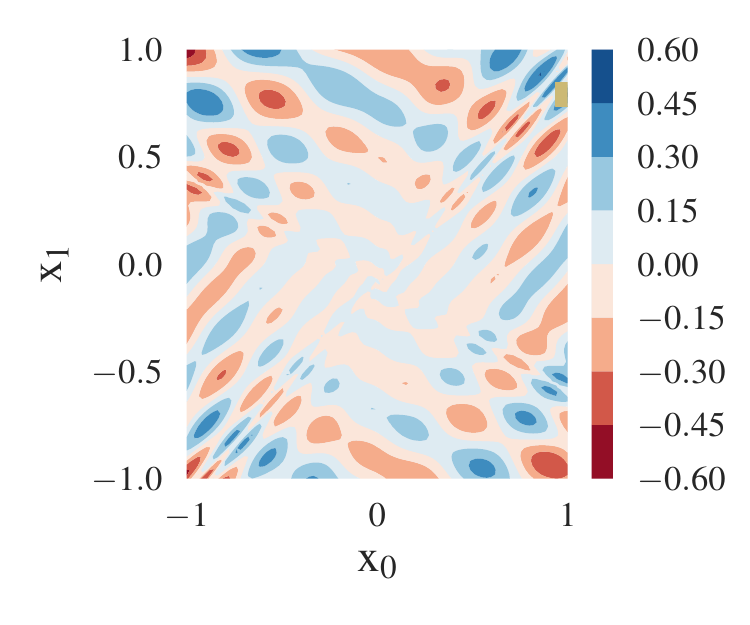}\\
	(a) Branin.
	&
	(b) Eggholder.
\end{tabular}
\caption{Synthetic functions whose maximizers are denoted as yellow squares.}
\label{fig:funcs}
\end{figure}

In this section, we consider synthetic function benchmarks: 
the Branin-Hoo function (Fig.~\ref{fig:funcs}a) and the eggholder function (Fig.~\ref{fig:funcs}b). 
The experiments are conducted with both small and large noise standard deviations: $\sigma_n = 0.01$ and $\sigma_n = 0.3$.
As the Branin-Hoo and eggholder functions are often used in minimization problems, the negative values of these functions are used as the objective function. Experiments with a function sample drawn from a GP and the $2$-dimensional Michaelwicz function are in the Appendix~\ref{app:otherexp}.

\begin{figure}[h!]
\centering
\begin{tabular}{@{}l@{}r@{}}
	\includegraphics[height=0.215\textwidth]{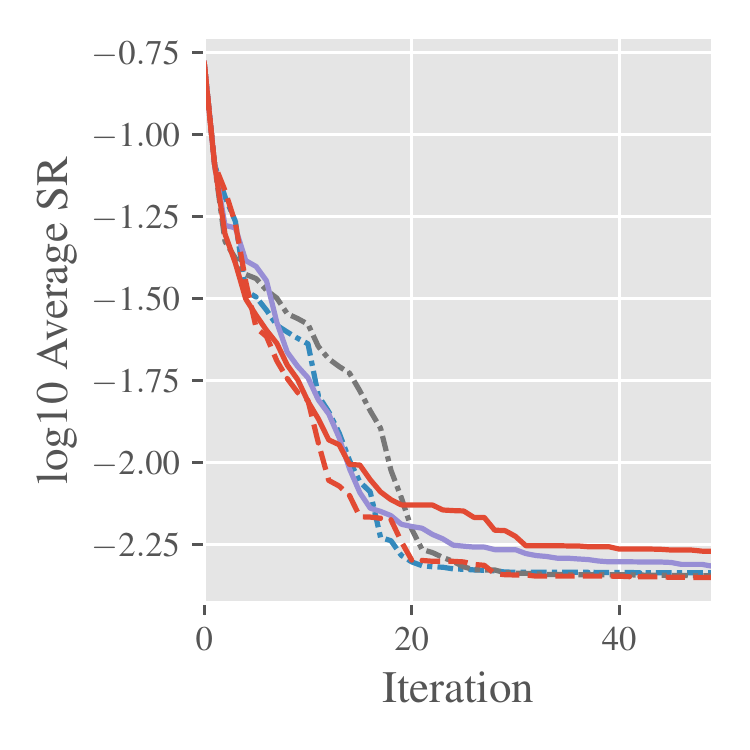}
	&
	\includegraphics[height=0.215\textwidth]{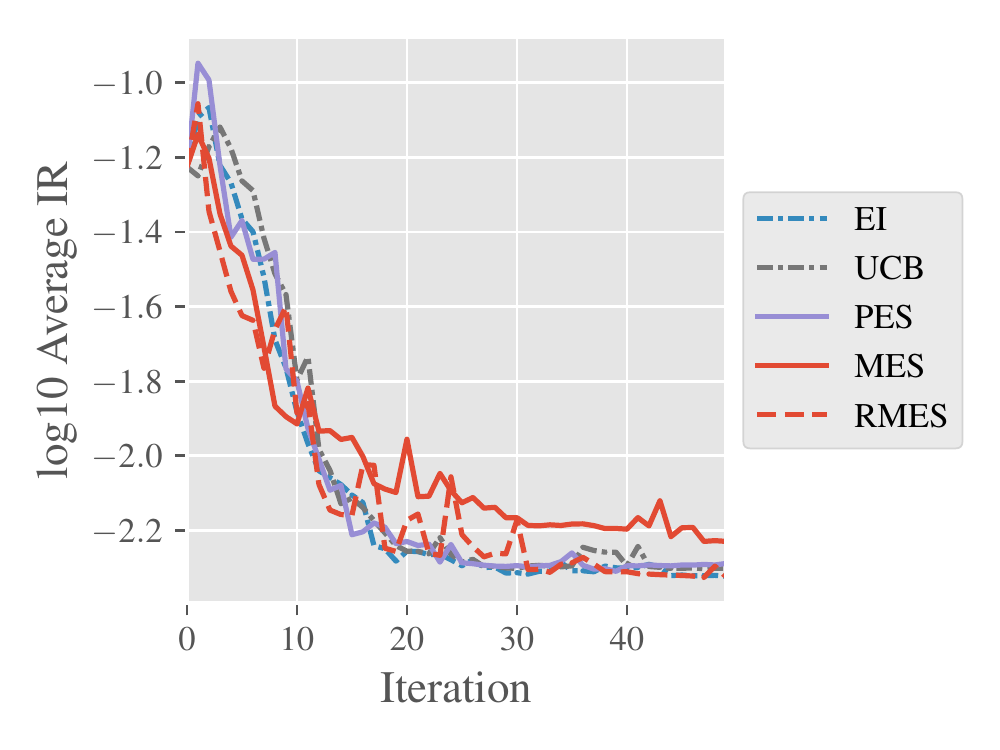}
	\\
	\multicolumn{2}{c}{(a) $\sigma_n = 0.01$.}
	\\
	\includegraphics[height=0.215\textwidth]{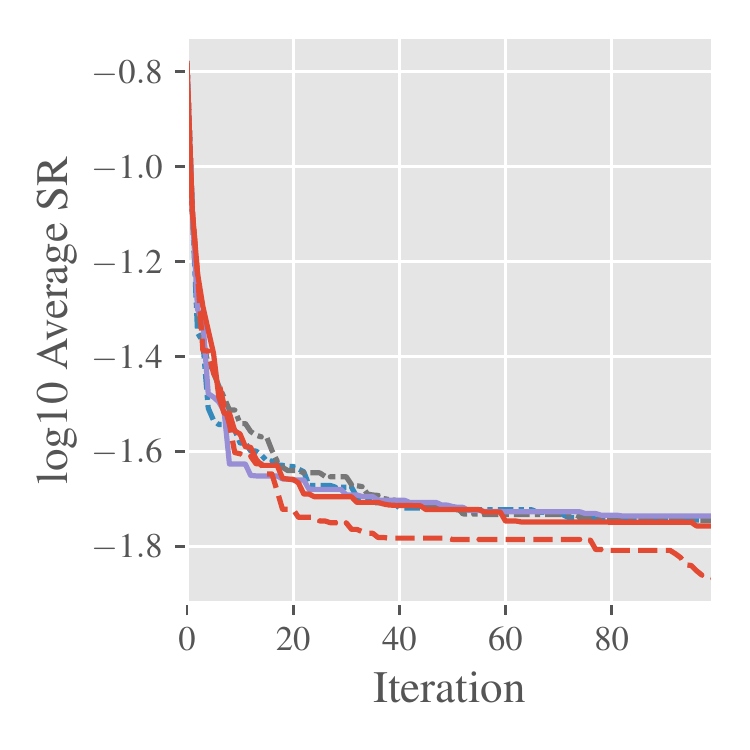}
	&
	\includegraphics[height=0.215\textwidth]{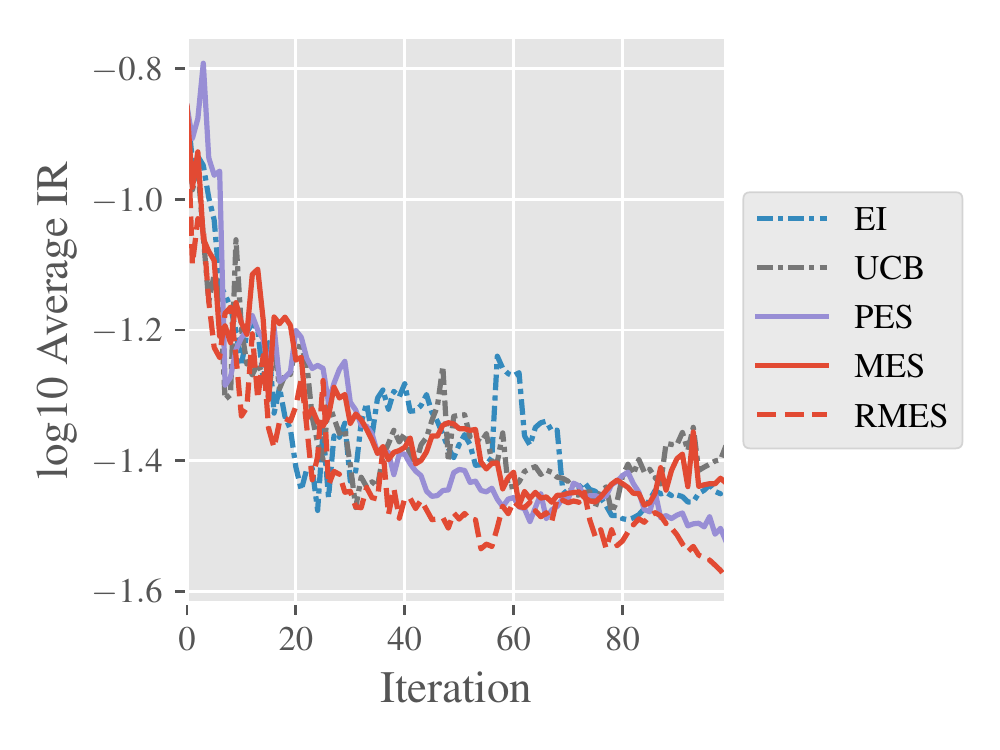}
	\\
	\multicolumn{2}{c}{(b) $\sigma_n = 0.3$.}
\end{tabular}
\caption{Branin-Hoo function.}
\label{fig:branin}
\end{figure}

\begin{figure}[h!]
\centering
\begin{tabular}{@{}l@{}r@{}}
	\includegraphics[height=0.215\textwidth]{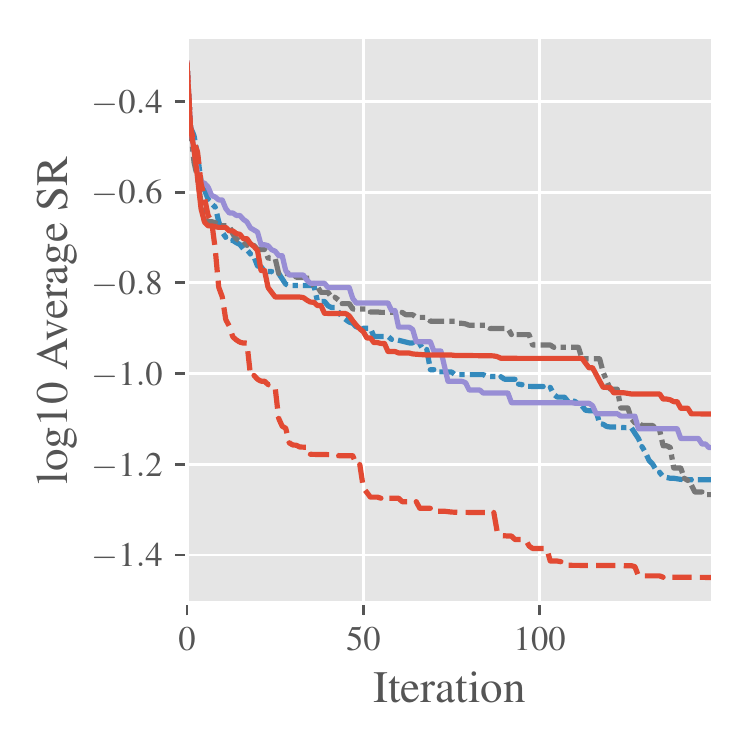}
	&
	\includegraphics[height=0.215\textwidth]{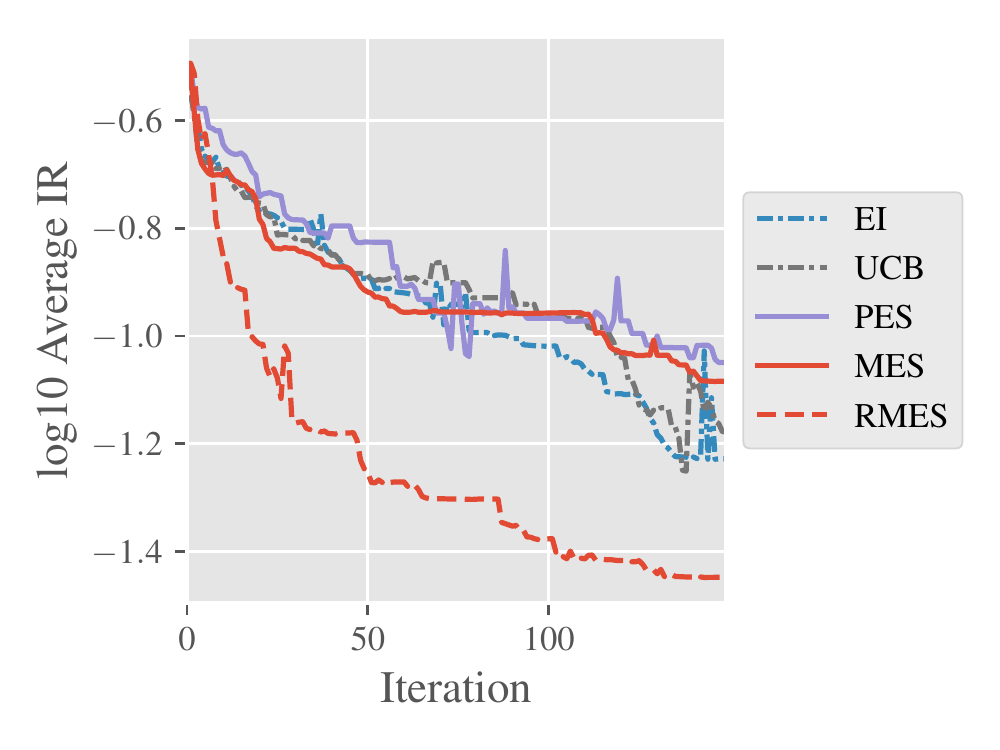}
	\\
	\multicolumn{2}{c}{(a) $\sigma_n = 0.01$.}
	\\
	\includegraphics[height=0.215\textwidth]{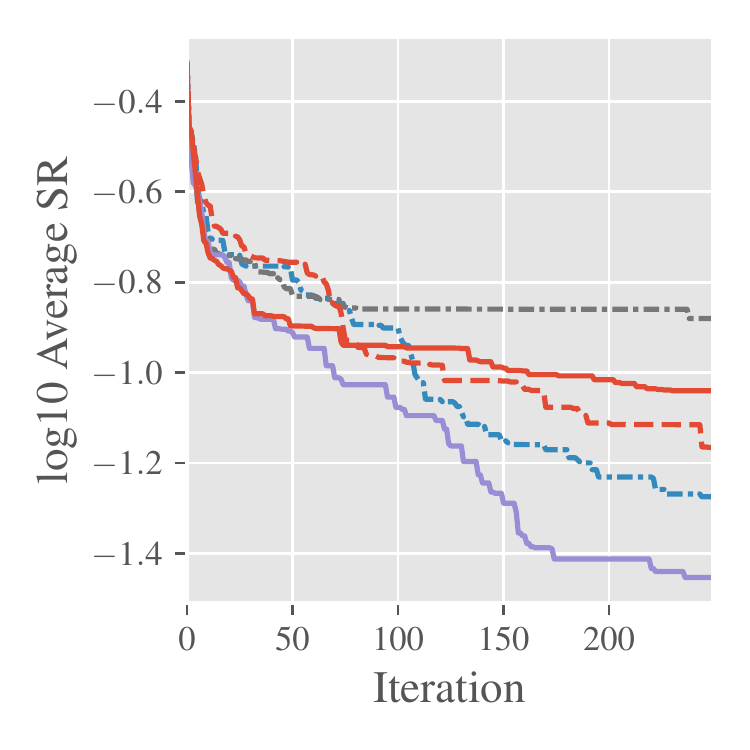}
	&
	\includegraphics[height=0.215\textwidth]{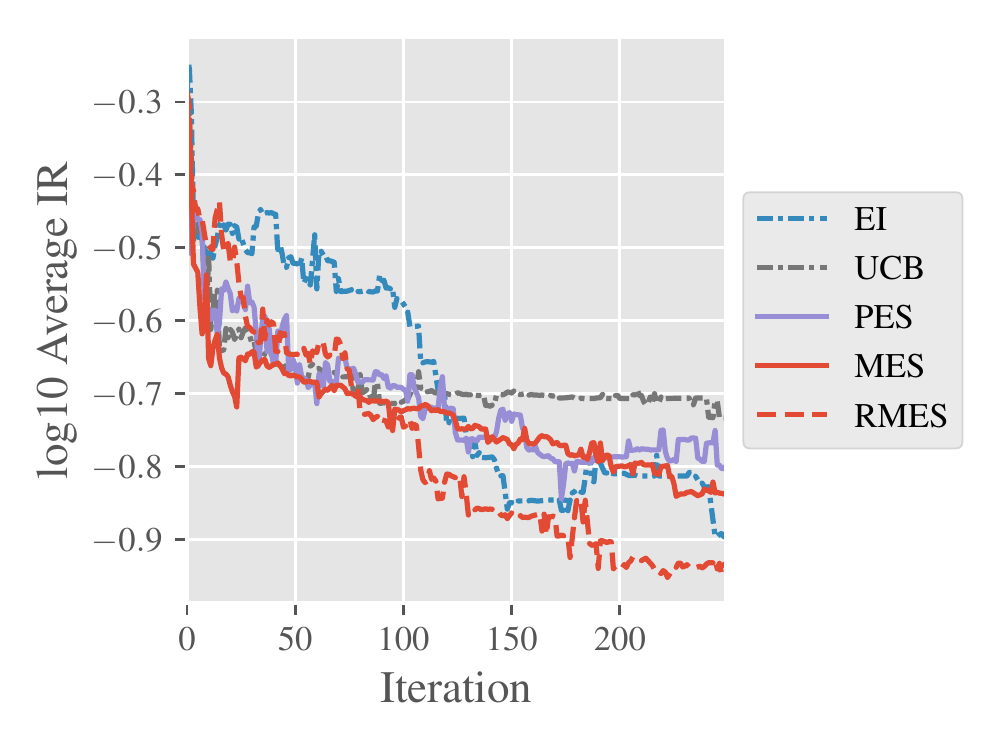}
	\\
	\multicolumn{2}{c}{(b) $\sigma_n = 0.3$.}
\end{tabular}
\caption{Eggholder function.}
\label{fig:eggholder}
\end{figure}

The logarithm to the base $10$ of the average of the simple regret (SR) and that of the inference regret (IR) at each BO iteration
are shown in Figs.~\ref{fig:branin} and~\ref{fig:eggholder}.
In general, RMES outperforms MES in all these experiments by converging to both small simple and inference regrets. This empirically illustrates the beneficial effects of correcting misconceptions in MES. In particular, when the noise is small, i.e., $\sigma_n = 0.01$, it is likely that the over-exploration of MES prevents it from properly exploiting the location of the maximizer as shown in Fig.~\ref{fig:comexplore}. On the other hand, when the noise is large, i.e., $\sigma_n = 0.3$, the over-exploitation traps MES at suboptimal maxima.

\begin{figure}[h!]
\centering
\begin{tabular}{@{}c@{}c@{}}
	\includegraphics[height=0.215\textwidth]{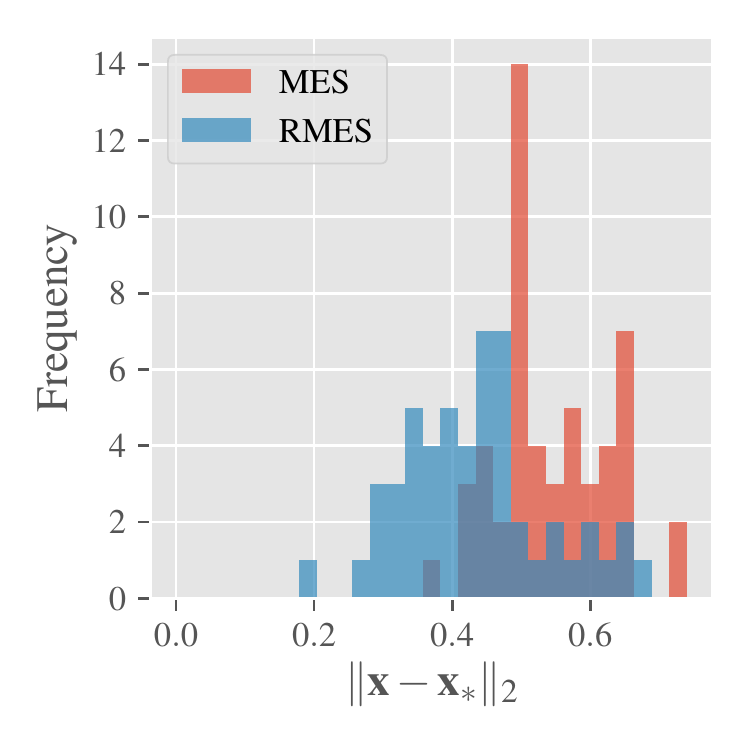}
	&
	\includegraphics[height=0.215\textwidth]{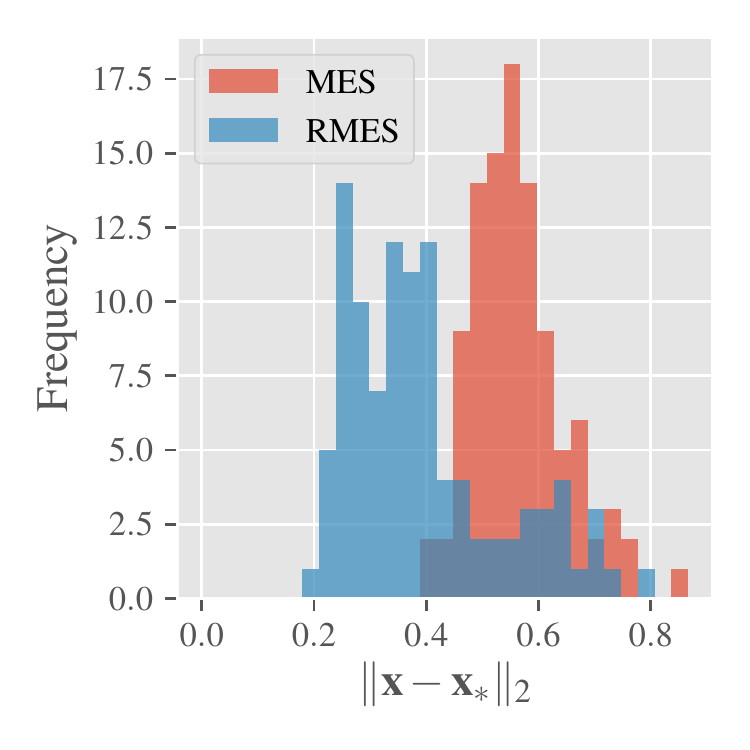}
	\\
	(a) $\sigma_n = 0.01$.
	&
	(b) $\sigma_n = 0.3$.
\end{tabular}
\caption{Distance of input queries to the maximizer of the Branin-Hoo function.}
\label{fig:branindist}
\end{figure}

The Branin-Hoo function is a simple function with a high correlation between function values in the input domain (Fig.~\ref{fig:funcs}a), so all the acquisition functions perform similarly in Fig.~\ref{fig:branin}. Nonetheless, MES still underperforms slightly in comparison with RMES. Fig.~\ref{fig:branindist} shows the histogram of the average distance from input queries to the maximizer over $15$ repetitions of the experiment, i.e., $\Vert \mbf{x} - \mbf{x}_*\Vert_2 \triangleq \sqrt{(\mbf{x} - \mbf{x}_*)^\top (\mbf{x} - \mbf{x}_*)}$, for RMES and MES. Due to the high correlation between function values, it does not require either RMES or MES to query for inputs close to the maximizer to obtain a good performance. However, we still observe that MES queries for inputs farther from the maximizer than RMES, which explains its poorer performance. 
This phenomenon is most likely due to the imbalance in the exploration-exploitation trade-off as illustrated in Section~\ref{sec:rmes}.
The same observation is noted in other experiments: the function sample drawn from a GP and the Michaelwicz function in Fig.~\ref{fig:f2dlgdist} and Fig.~\ref{fig:michaeldist} in the Appendix~\ref{app:otherexp}, respectively.

The eggholder function is a difficult function to optimize as it has many local maxima (Fig.~\ref{fig:funcs}b).
Hence, it requires an acquisition function to balance between exploration and exploitation to be query efficient. 
In this case, RMES outperforms other acquisition functions by converging to a smaller regret except for the simple regret when $\sigma_n = 0.3$ in Fig.~\ref{fig:eggholder}.

\subsection{Real-world Optimization Problems}

This section presents the results of BO algorithms for real-world optimization problems including an environment sensing problem of finding the location with the maximum pH value (Fig.~\ref{fig:funcs2}b in Appendix~\ref{app:otherexp}), and a machine learning training problem of tuning an SVM model for training the Wisconsin breast cancer dataset. The noise standard deviation values of the pH field and tuning the SVM model problems are $\sigma_n = 0.25$ and $\sigma_n = 0.02$, respectively. Although both MES and RMES show reasonable performance in these experiments, RMES has an advantage in the simple regret of the tuning SVM experiment. 
In comparison with EI and UCB, the difference in performance is insignificant.
Regarding PES, it does not perform as well as the other acquisition functions, especially in the tuning SVM experiment.

\begin{figure}[h!]
\centering
\begin{tabular}{@{}l@{}r@{}}
	\includegraphics[height=0.215\textwidth]{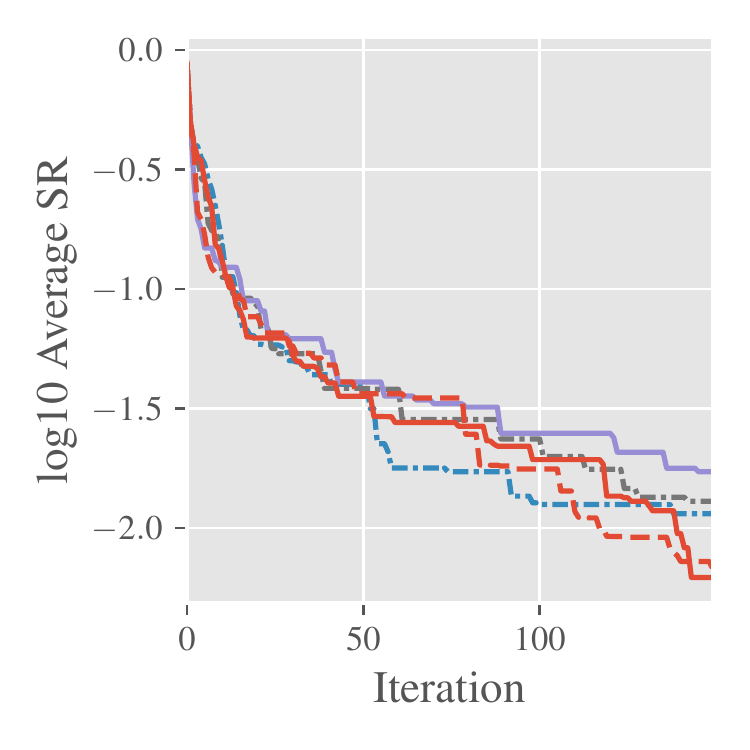}
	&
	\includegraphics[height=0.215\textwidth]{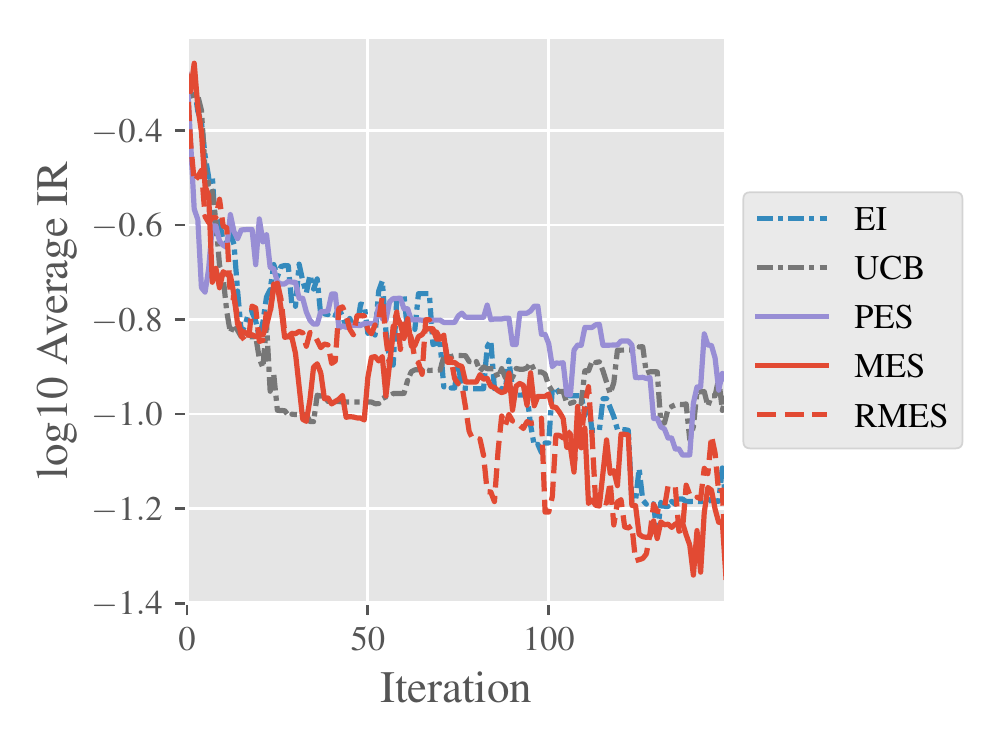}
\end{tabular}
\caption{The pH field experiment.}
\label{fig:pH}
\end{figure}

\begin{figure}[h!]
\centering
\begin{tabular}{@{}l@{}r@{}}
	\includegraphics[height=0.215\textwidth]{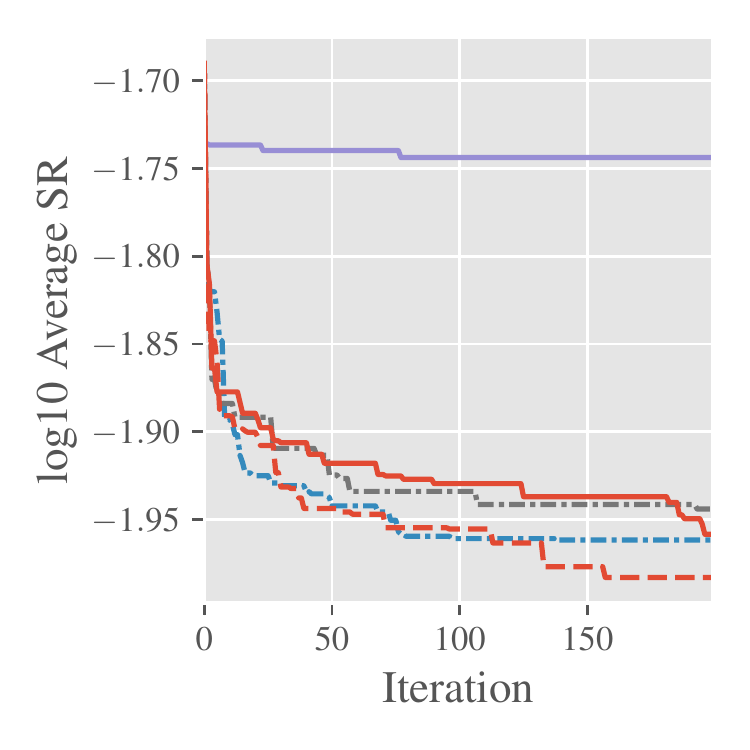}
	&
	\includegraphics[height=0.215\textwidth]{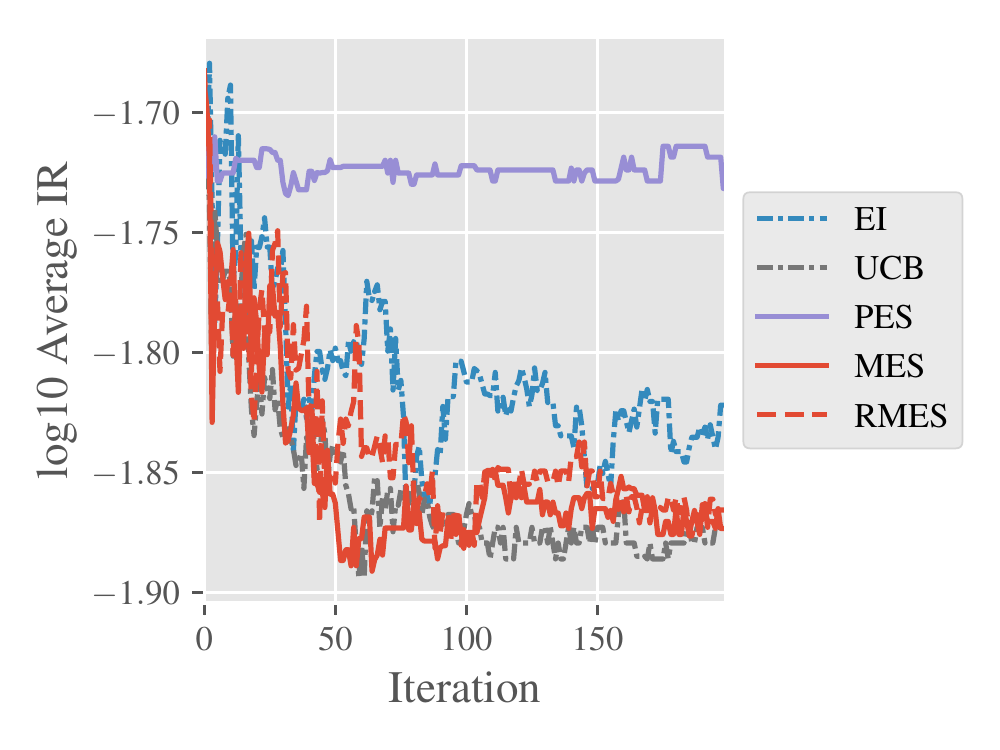}
\end{tabular}
\caption{Tuning SVM model experiment.}
\label{fig:svm}
\end{figure}

\section{Conclusion}

In this paper, we illustrate two misconceptions in the existing MES acquisition function: noiseless observations and discrepancy in the evaluation, which have negative implications on its interpretation as the mutual information as well as on its empirical performance. Based on the insights of these issues, we develop the RMES  acquisition function that produces a more accurate measure of the information gain about the max-value through observing noisy function outputs.
As a result, it has a superior performance compared with MES thanks to the correction of these issues. Nonetheless, optimizing RMES is more challenging than the existing MES as it does not have any closed-form expression. To overcome this hurdle, we derive a closed-form expression for the probability density of the noisy observation given the max-value, and design an efficient sampling approach to do stochastic gradient ascent with a reparameterization trick. 
It is empirically shown in several synthetic function benchmarks and real-world optimization problems that the performance of RMES is preferable over MES and competitive with existing acquisition functions.

\bibliography{bo}

\clearpage
\appendix

\section{Probability Density of $y_{\mbf{x}}|\mbf{y}_{\mcl{D}}, f_*$}
\label{app:pdfntn}

This section derives a closed-form expression of $p(y_{\mbf{x}}|\mbf{y}_{\mcl{D}}, f_*)$.
Different from the relatively straightforward expression in \eqref{eq:approxpy},
we can express $p(y_{\mbf{x}}|\mbf{y}_{\mcl{D}}, f_*)$ in an importance sampling manner. That is, samples of $y_{\mbf{x}}|\mbf{y}_{\mcl{D}}, f_*$ can be obtained by first drawing a sample $y$ of $y_{\mbf{x}}|\mbf{y}_{\mcl{D}}$ and then, weighting the sample with $p(f_{\mbf{x}} \le f_*|\mbf{y}_{\mcl{D}}, y_{\mbf{x}} = y)$.
Let $y$ denote a realization of the random variable $y_{\mbf{x}}$, we have
\begin{align}
&p(y_{\mbf{x}} = y|\mbf{y}_{\mcl{D}}, f_*)\nonumber\\
	&\propto p(y_{\mbf{x}} = y| \mbf{y}_{\mcl{D}}) p(f_{\mbf{x}} \le f_*| \mbf{y}_{\mcl{D}}, y_{\mbf{x}} = y)
\label{eq:theo1py}
\end{align}
where $p(f_{\mbf{x}} \le f_*| \mbf{y}_{\mcl{D}}, y_{\mbf{x}} = y)$ can be considered as the cumulative density function of the distribution specified by $p(f_{\mbf{x}}| \mbf{y}_{\mcl{D}}, y_{\mbf{x}} = y)$. 
Let $f$ denote a realization of the random variable $f_{\mbf{x}}$.
Recall that $y_{\mbf{x}} = f_{\mbf{x}} + \epsilon$ where the noise $\epsilon \sim \mcl{N}(0,\sigma_n^2)$,
we can evaluate the probability density $p(f_{\mbf{x}}=f| \mbf{y}_{\mcl{D}}, y_{\mbf{x}} = y)$ as below:
\begin{align*}
&p(f_{\mbf{x}} = f | \mbf{y}_{\mcl{D}}, y_{\mbf{x}} = y)\\
	&= p(f_{\mbf{x}} = f| \mbf{y}_{\mcl{D}}) p(\epsilon = y - f)\\
	&= \mcl{N}(f; \mu_{\mbf{x}}, \sigma_{\mbf{x}}^2) \mcl{N}(y-f; 0, \sigma_n^2)\\
	&= \mcl{N}\left(
		f;
		\frac{\sigma_n^2 \mu_{\mbf{x}} + \sigma_{\mbf{x}}^2 y}{\sigma_+^2},
		\frac{\sigma_{\mbf{x}}^2 \sigma_n^2}{\sigma_{+}^2}
	\right)\\
	&= \psi\left(
		\frac{\sigma_+^2 f - \sigma_n^2 \mu_{\mbf{x}} - \sigma_{\mbf{x}}^2 y}{\sigma_{\mbf{x}} \sigma_n \sigma_+}
	\right)
\end{align*}
where $ p(f_{\mbf{x}} = f| \mbf{y}_{\mcl{D}}) = \mcl{N}(f;\mu_{\mbf{x}}, \sigma_{\mbf{x}}^2)$ is the GP posterior distribution \eqref{eq:gppost}; $\psi$ denotes the probability density the standard Gaussian distribution, and $\sigma_+^2 \triangleq \sigma_{\mbf{x}}^2 + \sigma_n^2$.
Hence, we have the cumulative density function $p(f_{\mbf{x}} \le f_*| \mbf{y}_{\mcl{D}}, y_{\mbf{x}} = y)$ expressed as:
\begin{align}
p(f_{\mbf{x}} \le f_*| \mbf{y}_{\mcl{D}}, y_{\mbf{x}} = y)
	= \Psi\left(
		\frac{\sigma_+^2 f_* - \sigma_n^2 \mu_{\mbf{x}} - \sigma_{\mbf{x}}^2 y}{\sigma_{\mbf{x}} \sigma_n \sigma_+}
	\right)
\label{eq:cdffo}
\end{align}
where $\Psi$ denotes the cumulative density function of the standard Gaussian distribution. By substituting \eqref{eq:cdffo} into \eqref{eq:theo1py}, we obtain
\begin{align}
&p(y_{\mbf{x}} = y| \mbf{y}_{\mcl{D}}, f_*)\nonumber\\
	&\propto p(y_{\mbf{x}} = y| \mbf{y}_{\mcl{D}})
		\Psi\left(
		\frac{\sigma_+^2 f_* - \sigma_n^2 \mu_{\mbf{x}} - \sigma_{\mbf{x}}^2 y}{\sigma_{\mbf{x}} \sigma_n \sigma_+}
		\right)\nonumber\\
	&= \mcl{N}(y; \mu_{\mbf{x}}, \sigma_+^2)
		\Psi\left(
		\frac{\sigma_+^2 f_* - \sigma_n^2 \mu_{\mbf{x}} - \sigma_{\mbf{x}}^2 y}{\sigma_{\mbf{x}} \sigma_n \sigma_+}
		\right)\nonumber\\
	&= \mcl{N}(y; \mu_{\mbf{x}}, \sigma_+^2)
		\Psi\left(
			g_{f_*}(y)
		\right)
	\label{eq:theo1pyexpr}
\end{align}
where $\displaystyle g_{f_*}(y) \triangleq \frac{\sigma_+^2 f_* - \sigma_n^2 \mu_{\mbf{x}} - \sigma_{\mbf{x}}^2 y}{\sigma_{\mbf{x}} \sigma_n \sigma_+}$.
To obtain the expression for $p(y_{\mbf{x}} = y| \mbf{y}_{\mcl{D}}, f_*)$, we need to evaluate the integral of \eqref{eq:theo1pyexpr}:
\begin{align}
&\int \mcl{N}(y; \mu_{\mbf{x}}, \sigma_+^2) \Psi\left( g_{f_*}(y) \right)\ \text{d}y\nonumber\\
&= \int p(y_{\mbf{x}} = y|\mbf{y}_{\mcl{D}}) \Psi\left(
		\frac{\sigma_+^2 f_* - \sigma_n^2 \mu_{\mbf{x}} - \sigma_{\mbf{x}}^2 y}{\sigma_{\mbf{x}} \sigma_n \sigma_+}
		\right)\ \text{d}y\nonumber\\
&= \int p(y_{\mbf{x}} = y|\mbf{y}_{\mcl{D}}) p\left(
		\nu \le 
		\frac{\sigma_+^2 f_* - \sigma_n^2 \mu_{\mbf{x}} - \sigma_{\mbf{x}}^2 y}{\sigma_{\mbf{x}} \sigma_n \sigma_+}
		\right)\ \text{d}y\nonumber\\
&= p\left(
		\nu \le 
		\frac{\sigma_+^2 f_* - \sigma_n^2 \mu_{\mbf{x}} - \sigma_{\mbf{x}}^2 y_{\mbf{x}}}{\sigma_{\mbf{x}} \sigma_n \sigma_+} \Big| \mbf{y}_{\mcl{D}}
		\right)\nonumber\\
&= p\left(
	\nu 	\sigma_{\mbf{x}} \sigma_n \sigma_+ + \sigma_{\mbf{x}}^2 y_{\mbf{x}} \le \sigma_+^2 f_* - \sigma_n^2 \mu_{\mbf{x}} | \mbf{y}_{\mcl{D}}
	\right)\nonumber
\end{align}
where $\nu \sim \mcl{N}(0,1)$. Recall that $y_{\mbf{x}} | \mbf{y}_{\mcl{D}} \sim \mcl{N}(\mu_{\mbf{x}}, \sigma_+^2)$, it implies $\nu 	\sigma_{\mbf{x}} \sigma_n \sigma_+ + \sigma_{\mbf{x}}^2 y_{\mbf{x}}$ follows a Gaussian distribution $\mcl{N}(\sigma_{\mbf{x}}^2 \mu_{\mbf{x}}, \sigma_{\mbf{x}}^2 \sigma_+^4)$. Therefore, $p(
	\nu \sigma_{\mbf{x}} \sigma_n \sigma_+ + \sigma_{\mbf{x}}^2 y_{\mbf{x}} \le \sigma_+^2 f_* - \sigma_n^2 \mu_{\mbf{x}} | \mbf{y}_{\mcl{D}})$ is the cumulative density function at $\sigma_+^2 f_* - \sigma_n^2 \mu_{\mbf{x}}$ of a Gaussian distribution $\mcl{N}(\sigma_{\mbf{x}}^2 \mu_{\mbf{x}}, \sigma_{\mbf{x}}^2 \sigma_+^4)$, i.e.,
\begin{align}
&\int \mcl{N}(y; \mu_{\mbf{x}}, \sigma_+^2) \Psi\left( g_{f_*}(y) \right)\ \text{d}y\nonumber\\
&= \Psi\left( 
	\sigma_+^2 f_* - \sigma_n^2 \mu_{\mbf{x}};
	\sigma_{\mbf{x}}^2 \mu_{\mbf{x}},
	\sigma_{\mbf{x}}^2 \sigma_+^4
	\right)\nonumber\\
&= \Psi \left( \frac{f_* - \mu_{\mbf{x}}}{\sigma_{\mbf{x}}} \right)\ .
\label{eq:pynorm}
\end{align}
Hence, from \eqref{eq:theo1pyexpr} and \eqref{eq:pynorm}, we obtain the exact probability density function of $y_{\mbf{x}}|\mbf{y}_{\mcl{D}},f_*$ as below:
\begin{align}
p(y_{\mbf{x}}|\mbf{y}_{\mcl{D}},f_*) = \mcl{N}(y_{\mbf{x}}; \mu_{\mbf{x}}, \sigma_{+}^2) 
	\frac{
	\Psi(g_{f_*}(y_{\mbf{x}}))
	}
	{
	\Psi(h_{f_*}(\mbf{x}))
	}
\end{align}
where $\displaystyle g_{f_*}(y_{\mbf{x}}) \triangleq \frac{\sigma_+^2 f_* - \sigma_n^2 \mu_{\mbf{x}} - \sigma_{\mbf{x}}^2 y_{\mbf{x}}}{\sigma_{\mbf{x}} \sigma_n \sigma_+}$ and 
$\displaystyle h_{f_*}(\mbf{x}) \triangleq \frac{f_* - \mu_{\mbf{x}}}{\sigma_{\mbf{x}}} $.

\section{Other Synthetic Function Benchmarks}
\label{app:otherexp}

\begin{figure}[h!]
\centering
\begin{tabular}{@{}c@{}c@{}}
	\includegraphics[height=0.18\textwidth]{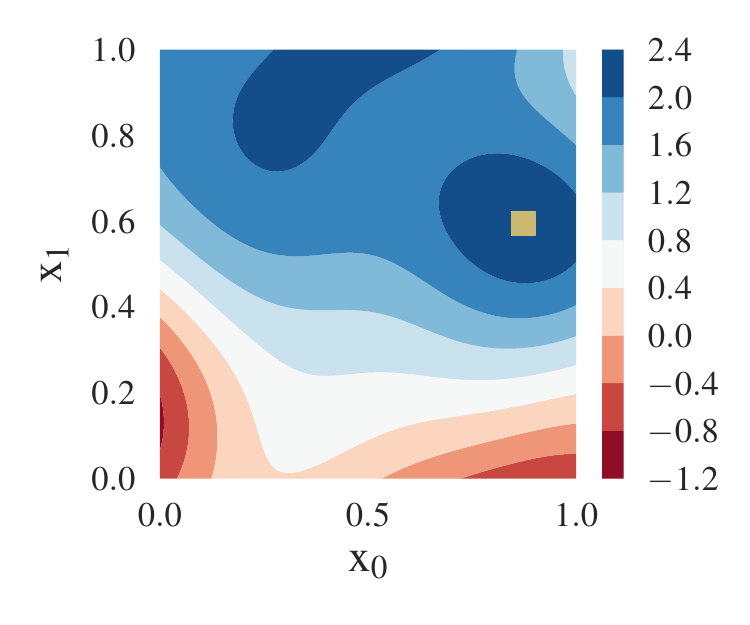}
	&
	\includegraphics[height=0.18\textwidth]{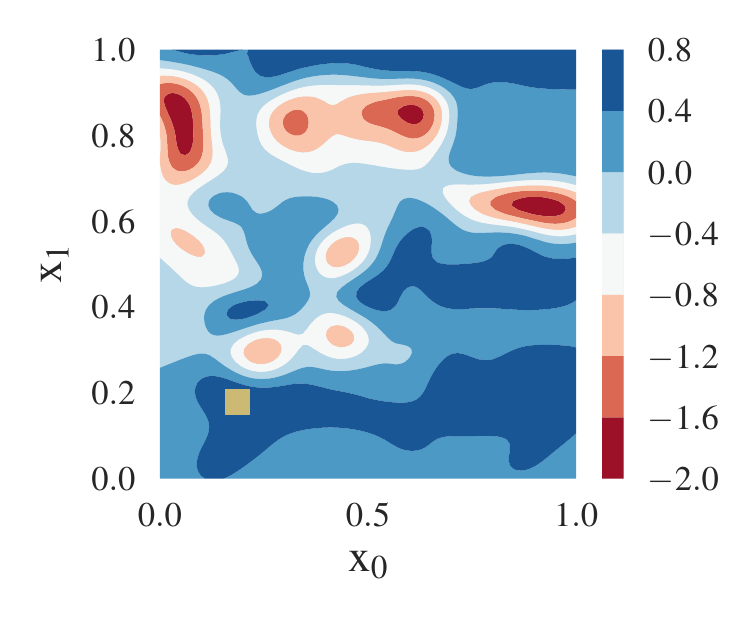}
	\\
	(a)
	&
	(b)
\end{tabular}
\caption{Function sample drawn from GP (a) and the GP posterior mean of the pH field (b).}
\label{fig:funcs2}
\end{figure}

In this section, we describe experiments with other synthetic function benchmarks including: a function sample drawn from a GP with hyperparameters: $l=0.33$, $\sigma_s^2=1$ (Fig.~\ref{fig:funcs2}a); and the $2$-dimensional Michaelwicz function. 

\begin{figure}[h!]
\centering
\begin{tabular}{@{}l@{}r@{}}
	\includegraphics[height=0.215\textwidth]{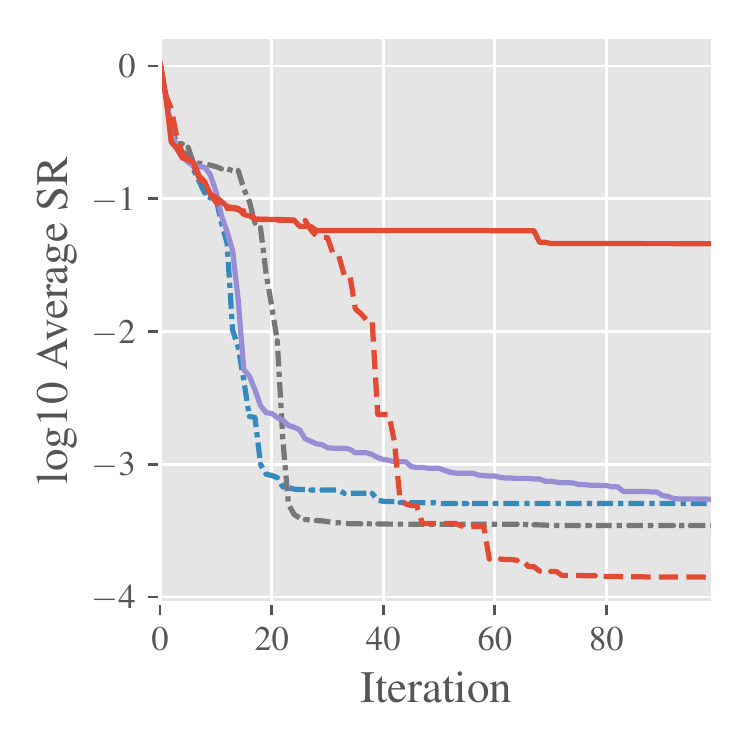}
	&
	\includegraphics[height=0.215\textwidth]{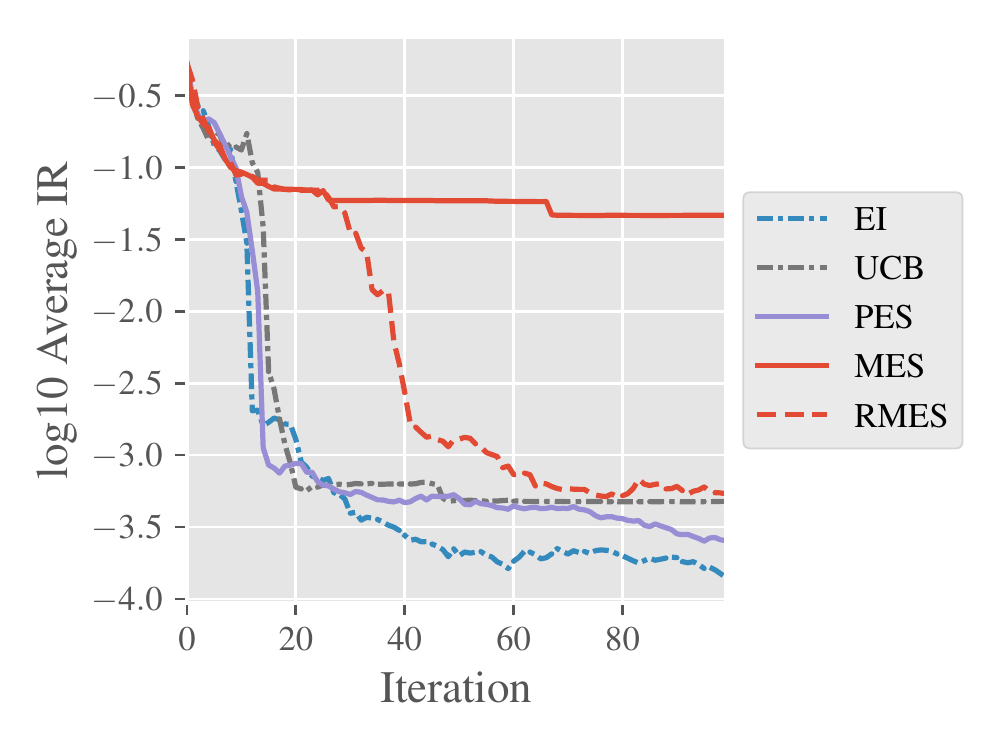}
	\\
	\multicolumn{2}{c}{(a) $\sigma_n = 0.01$.}
	\\
	\includegraphics[height=0.215\textwidth]{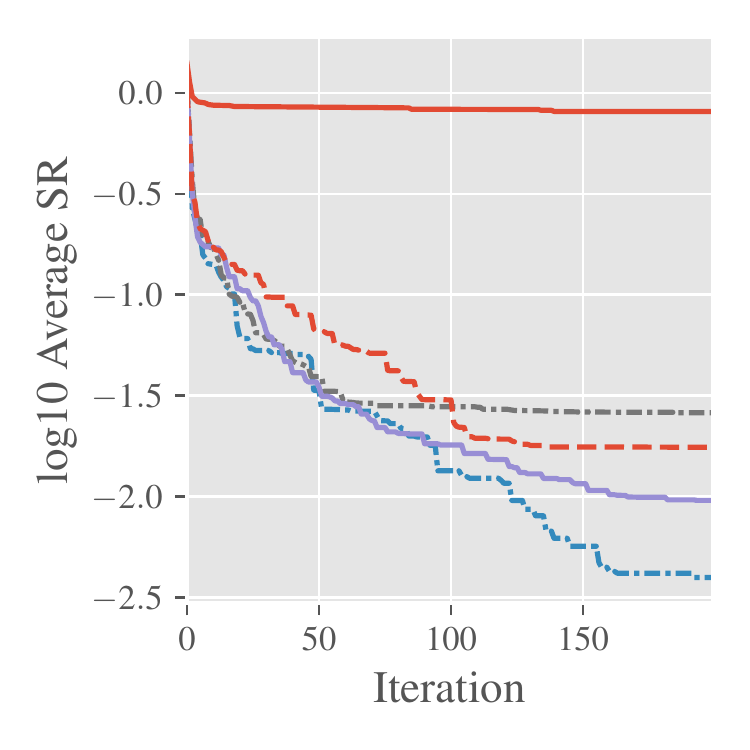}
	&
	\includegraphics[height=0.215\textwidth]{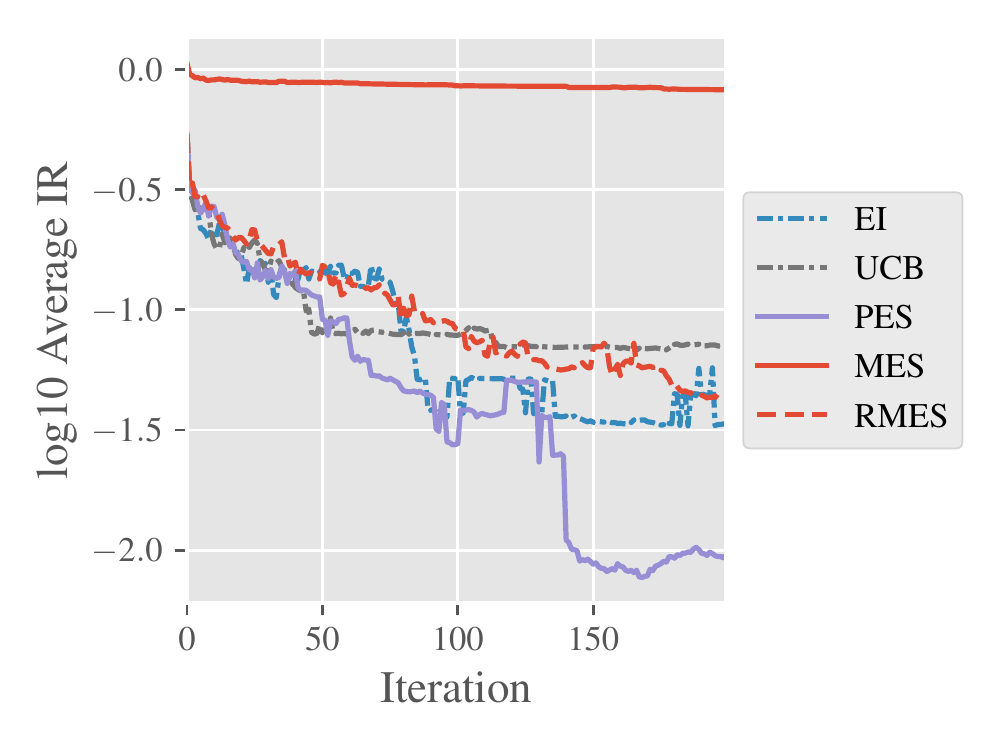}
	\\
	\multicolumn{2}{c}{(b) $\sigma_n = 0.3$.}
\end{tabular}
\caption{A function sample drawn from a GP.}
\label{fig:f2dlg}
\end{figure}

\begin{figure}[h!]
\centering
\begin{tabular}{@{}c@{}c@{}}
	\includegraphics[height=0.215\textwidth]{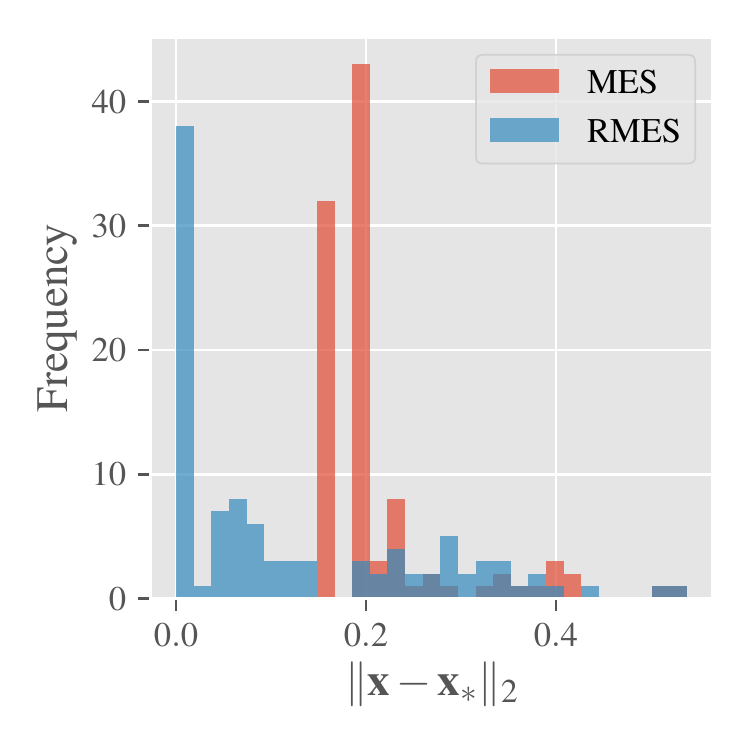}
	&
	\includegraphics[height=0.215\textwidth]{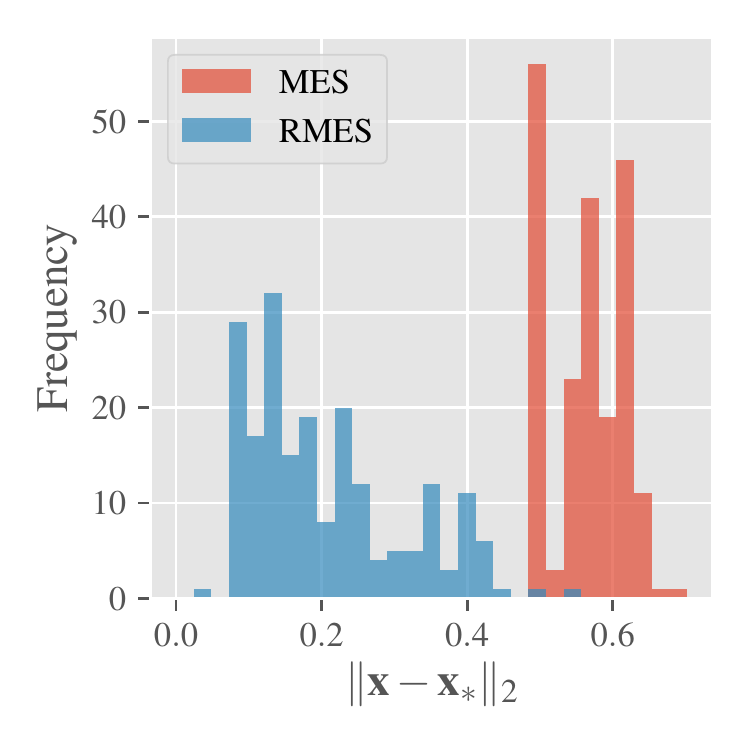}
	\\
	(a) $\sigma_n = 0.01$.
	&
	(b) $\sigma_n = 0.3$.
\end{tabular}
\caption{Distance of input queries to the maximizer of a function sample drawn from a GP.}
\label{fig:f2dlgdist}
\end{figure}

In the function sample drawn from a GP experiment (Fig.~\ref{fig:f2dlg}), MES converges to a much larger regret in comparison to other acquisition functions. We plot the distance from the input queries to the maximizer, i.e., $\Vert \mbf{x} - \mbf{x}_*\Vert_2 \triangleq \sqrt{(\mbf{x} - \mbf{x}_*)^\top (\mbf{x} - \mbf{x}_*)}$, for RMES and MES in Fig.~\ref{fig:f2dlgdist} which shows that MES does not query inputs close the maximizer for both values of $\sigma_n$ due to the imbalance in the exploration-exploitation trade-off.
On the other hand, RMES spends a large proportion of queries for inputs close to the maximizer.
RMES explores more than EI, UCB, and PES in this experiment as RMES converges slower in Fig.~\ref{fig:f2dlg}. However, RMES converges to a better simple regret when $\sigma_n=0.01$. As this function is relatively easy to optimize (Fig.~\ref{fig:funcs2}a), EI can quickly exploit to get to the maximizer, so it outperforms the other acquisition functions in the inference regret when $\sigma_n = 0.01$ and in the simple regret when $\sigma_n = 0.3$. On the other hand, PES achieves the best inference regret when $\sigma_n = 0.3$.

\begin{figure}[h!]
\centering
\begin{tabular}{@{}l@{}r@{}}
	\includegraphics[height=0.215\textwidth]{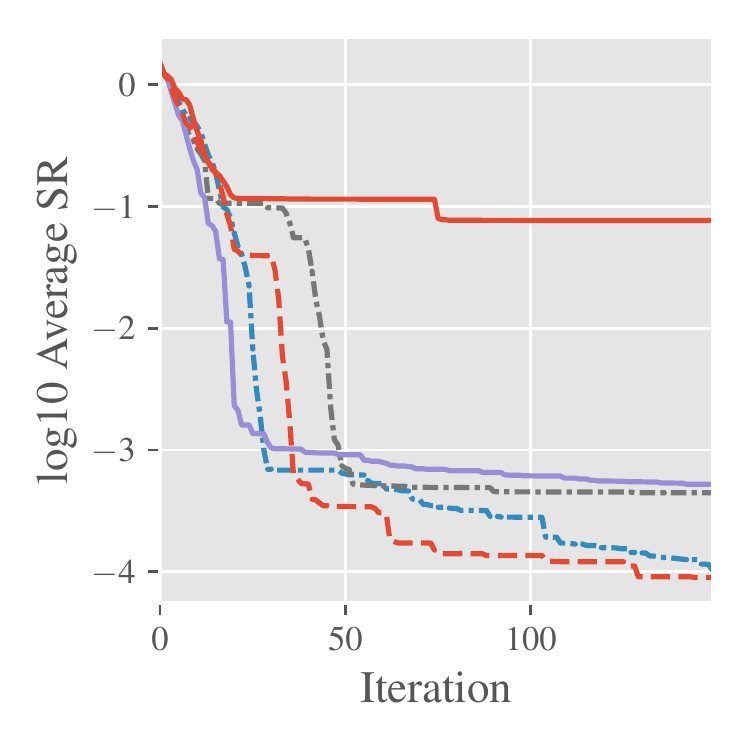}
	&
	\includegraphics[height=0.215\textwidth]{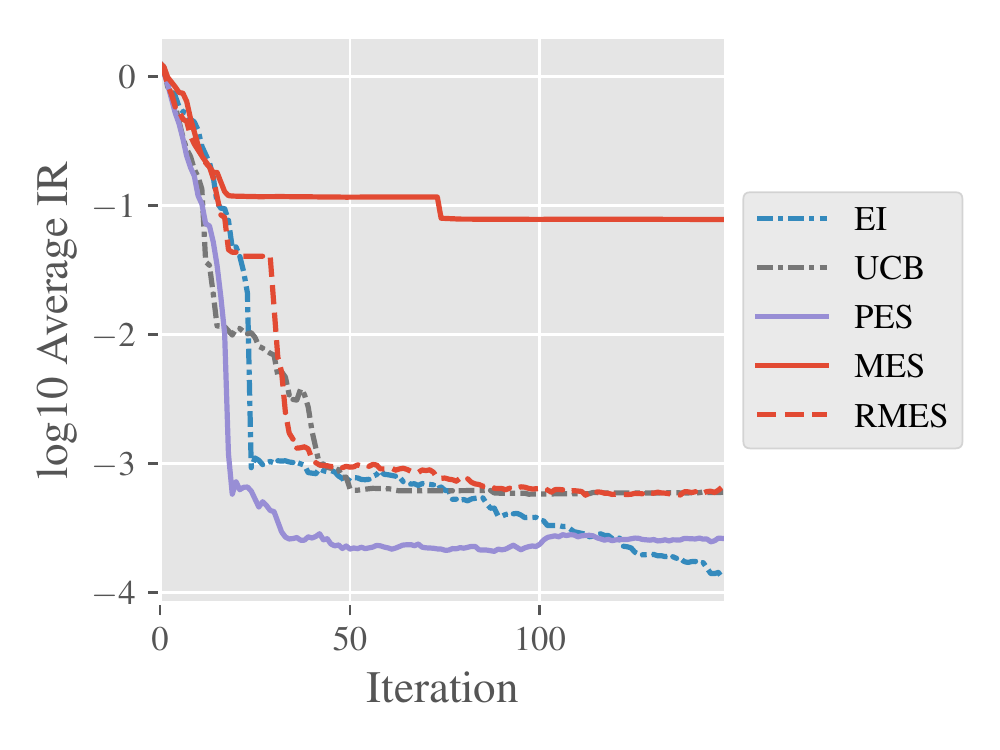}
\end{tabular}
\caption{Michaelwicz function with $\sigma_n = 0.01$.}
\label{fig:michael}
\end{figure}

\begin{figure}[h!]
\centering
	\includegraphics[height=0.215\textwidth]{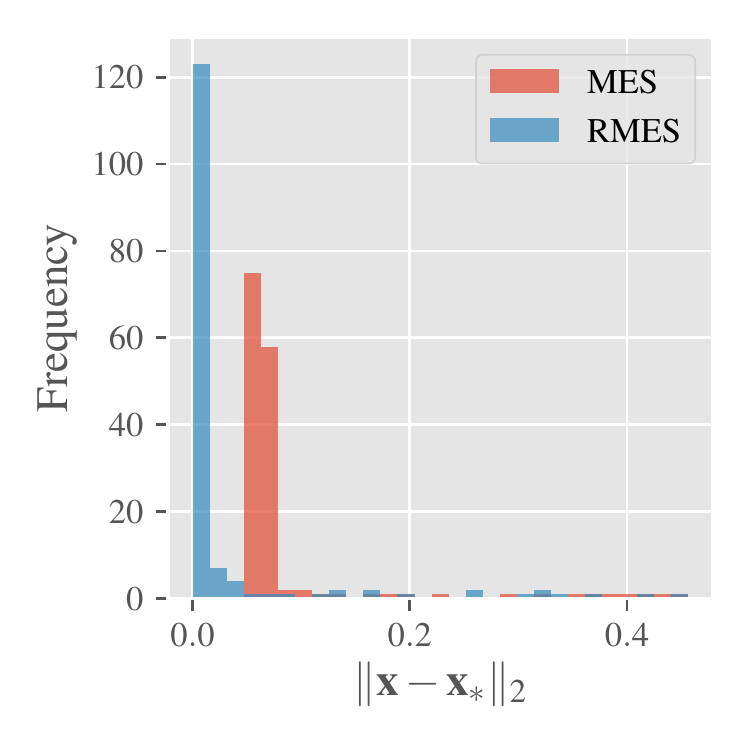}
\caption{Distance of input queries to the maximizer of the Michaelwicz function.}
\label{fig:michaeldist}
\end{figure}

Fig.~\ref{fig:michael} shows the results of the $2$-dimensional Michaelwicz function. 
We can observe that MES does not perform as well as the other acquisition functions.
Fig.~\ref{fig:michaeldist} of the distance of input queries to the maximizer also shows that MES does not query inputs close to the maximizer in comparison to RMES, which means MES cannot properly search for the maximizer.
Among EI, UCB, PES, and RMES, we observe that in terms of the simple regret, RMES is on par with EI and they outperform the other acquisition functions. Regarding the inference regret, PES and EI have the best performance though RMES matches the performance of UCB.

\end{document}